\documentclass[runningheads]{llncs}

\usepackage{eccv}

\usepackage{eccvabbrv}
\usepackage{graphicx}
\usepackage{booktabs}
\usepackage[accsupp]{axessibility}
\usepackage{wrapfig}
\usepackage{caption}
\usepackage{algorithm}
\usepackage{algpseudocode}
\usepackage{tabularx}
\usepackage{hyperref}
\usepackage{orcidlink}

\usepackage{multirow}
\usepackage{booktabs}
\usepackage{amsmath}
\usepackage{graphicx}
\usepackage{cuted}
\usepackage{capt-of}
\usepackage{algorithm}
\usepackage{algpseudocode}
\usepackage{etoc}
\usepackage{gensymb}
\usepackage{makecell}
\usepackage{float}

\def\ie{\emph{i.e.}\xspace}

\newcommand{\Fig}[1]{Fig.\ref{fig:#1}}

\newcommand{\Alg}[1]{Alg.\ref{alg:#1}}

\newcommand{\Tbl}[1]{Tab.\ref{tab:#1}}
\newcommand{\Tblstart}[1]{Table~\ref{tab:#1}}
\newcommand{\Figstart}[1]{Figure~\ref{fig:#1}}

\begin{document}
\hypersetup{hidelinks}

\title{WristMimic: Full-Body Humanoid Control with Wrist-Guided Manipulation}
\titlerunning{WristMimic}

\author{
Wongyun Yu\inst{1}\orcidlink{0009-0002-0161-8753} \and
Youngwoon Kim\inst{1}\orcidlink{0009-0000-6960-8338} \and
Minsu Cho\inst{1,2}\orcidlink{0000-0001-7030-1958}
}
\authorrunning{W.~Yu et al.}
\institute{
Pohang University of Science and Technology (POSTECH)\\
\and
RLWRLD\\
\url{https://wongyun-yu.github.io/wristmimic/}
}

\maketitle

\begin{abstract}
Retargeting human–object interaction demonstrations to \\physics-based simulation requires reproducing not only body motion but also the object motion and contacts that make manipulation succeed. However, position-only hand trajectories do not specify the contact forces needed to manipulate objects, and directly tracking them can overconstrain contact-rich finger behavior. We introduce WristMimic, a wrist-guided whole-body control framework that explicitly separates contact-free body motion from contact-rich hand manipulation. The contact-free body and wrist are guided by kinematic pose targets, whereas the fingers are not directly supervised by human hand pose. Instead, they learn grasping and manipulation behaviors from object tracking and contact outcomes. Our key insight is that the wrist is the natural gate between these two regimes. It is largely free from contact and can be tracked kinematically, yet it determines the global hand configuration and places the fingers within reachable grasp affordances. To ensure reliable wrist placement during interaction, we introduce wrist-specific reset constraints and reward prioritization. Experiments show that WristMimic matches or surpasses methods using full finger pose supervision while enabling finger-agnostic retargeting across diverse hand embodiments.
\keywords{ \ Whole body control \and Humanoid manipulation \and Physics-based retargeting }
\end{abstract}
\section{Introduction}
\label{sec:intro}
\begin{figure}[t]
    \centering
    \includegraphics[width=\columnwidth]{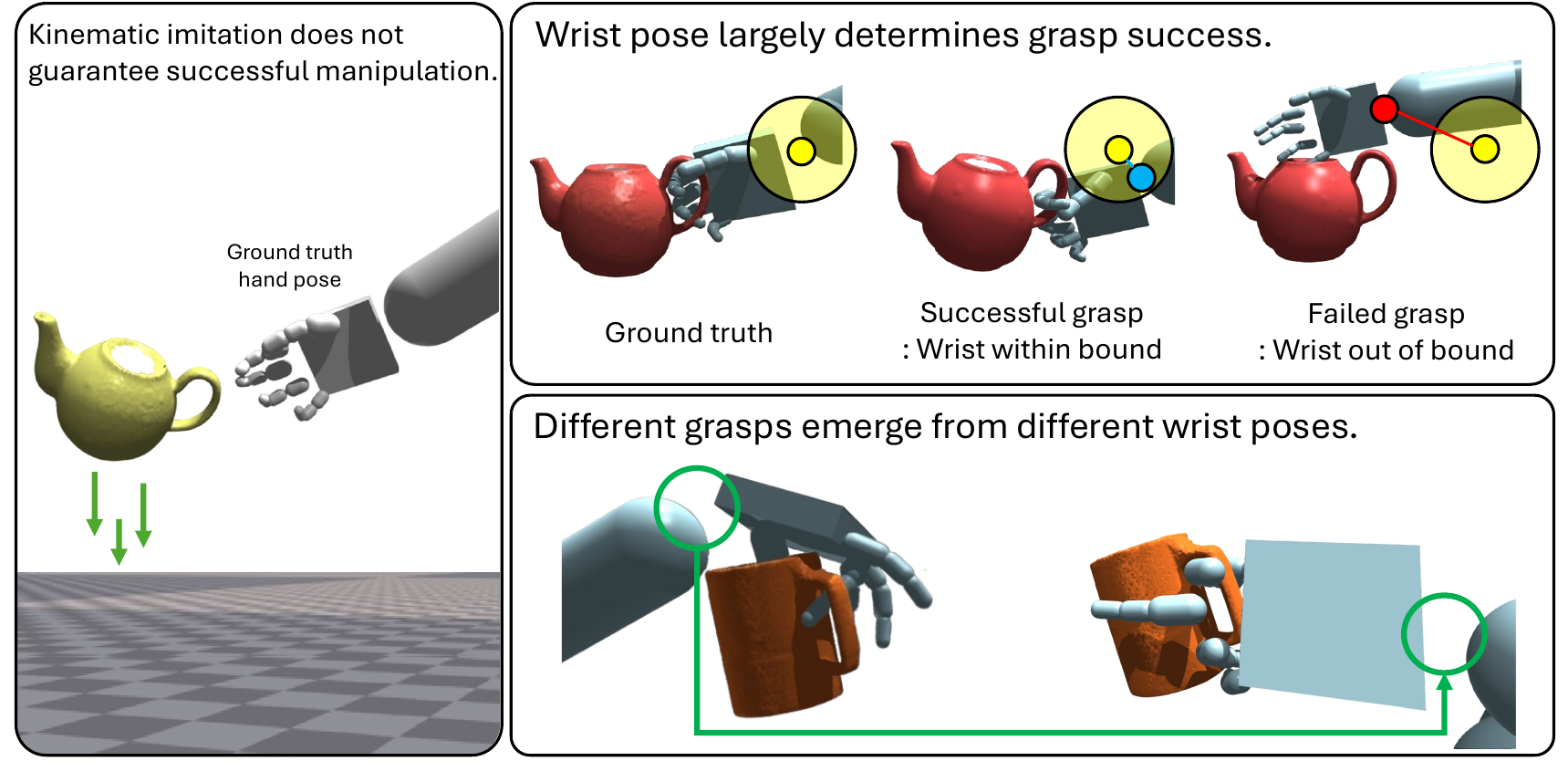}
\caption{\textbf{Kinematic imitation does not guarantee successful manipulation and wrist pose encodes grasp affordance.}
\textbf{Left:} Even with accurate hand pose imitation, manipulation may fail because interaction forces are not encoded in kinematic trajectories.
\textbf{Top right:} Wrist pose determines grasp success. Ground truth (left), feasible wrist region leading to stable grasp (middle, blue), and infeasible wrist pose causing failure (right, red). The yellow circle illustrates the feasible wrist boundary motivating our reset strategy.
\textbf{Bottom right:} Different wrist poses induce distinct grasp configurations, highlighting that wrist pose captures grasp affordance and enables finger-agnostic, wrist-guided manipulation.}
    \label{fig:teaser}
\end{figure}
Retargeting human-object interaction demonstrations to physics-based simulation is a long-standing problem in robotics and computer vision~\cite{omnih2o, humanplus, okami, expressive, hermes, vividex}. The goal of such retargeting extends beyond replicating body poses. Successful manipulation requires recovering correct object kinematics and contact dynamics in the target environment~\cite{chen2024object}. A common approach tracks kinematic trajectories from human demonstrations and replays them in simulation, which is effective for body parts that do not directly interact with the manipulated object, such as the torso, arms during reaching, and the wrist. However, for the hands, where most object interaction occurs, simply replaying kinematic trajectories is insufficient (\Fig{teaser}, left). For hands, the objective is not merely to reproduce human finger poses, but to reproduce the object motion and contacts that realize manipulation. However, position-based kinematic data alone cannot encode the contact forces that govern such manipulation. This observation raises a natural question: rather than pursuing increasingly precise finger tracking, can we let the hands be guided by object and contact dynamics instead?

We address this question by decoupling whole-body control into two regimes: kinematic pose-guided control for contact-free body parts and object- and contact-guided control for contact-rich hand manipulation. This decoupling reflects a fundamental asymmetry in physics-based interaction. Body motion without direct object contact is often well-described in position space, while finger motion during interaction is dominated by contact force dynamics that kinematic trajectories cannot capture. Instead of relying on finger kinematic references, we use object kinematics as an outcome-level cue of successful control, together with contact objectives to shape physical interaction.

The key component that makes this decoupling possible is the wrist. The wrist is largely free from object contact, making it amenable to kinematic guidance. At the same time, it determines the global pose of the fingers and plays a key role in both affordance and grasp success. Consider grasping a cup by its handle versus its top (\Fig{teaser}, bottom right): the wrist position and orientation differ significantly between these two affordances, while finger configurations can be learned from object-pose objectives and contact dynamics once the wrist is appropriately positioned.

Building on this wrist-centric insight, we introduce \textbf{WristMimic}, a whole-body control framework for affordance-aware manipulation that requires only body and wrist supervision. WristMimic positions the wrist in affordance-aware poses through wrist-specific reset boundaries and reward design, while allowing reinforcement learning to discover finger configurations through physics-based exploration, object-pose objectives, and contact dynamics without explicit finger supervision. As a consequence, this approach avoids relying on explicit finger pose supervision or complex finger-level capture data~\cite{interhand2, meng20223d, hot3d, arctic, oakink, oakink2, grab, freihand, assemblyhands, dexycb, h2o, ho3d}. Through extensive experiments, we demonstrate that precise wrist control achieves performance comparable to existing methods without relying on finger kinematic references, while supporting finger-agnostic retargeting across different hand embodiments.
Our main contributions are:
\begin{itemize}
\item We propose a decoupled whole-body control framework that separates kinematic pose-guided body control from object- and contact-guided finger manipulation for human-object interaction retargeting.
\item We identify the wrist as a critical bridge between control of contact-free body parts and contact-rich hand manipulation, and introduce wrist-specific reset boundaries and reward design to enable affordance-aware manipulation.
\item We demonstrate that precise wrist control without finger kinematics achieves performance comparable to methods using full finger pose supervision, while supporting finger-agnostic retargeting across different hand embodiments.
\end{itemize}

\section{Related Work}
\label{sec:related}
\subsection{Physics-based Motion Imitation}
\noindent{\textbf{Locomotion.}}
Physics-based motion imitation has progressed from single-clip tracking~\cite{deepmimic} to large-scale whole-body control. AMP~\cite{amp} and ASE~\cite{ase} introduced adversarial motion priors for stylized control without explicit reference tracking. Scaling to large motion datasets, ScaDiver~\cite{scadiver} employed mixture of experts, while PHC~\cite{phc} achieved high-fidelity imitation on AMASS~\cite{amass} dataset without external forces. PULSE~\cite{pulse} further distilled these motor skills into a universal latent representation reusable across downstream tasks. Recent universal trackers~\cite{gmt, beyondmimic, any2track, unitracker} further improved generalization to unseen motions, and SONIC~\cite{sonic} demonstrated favorable scaling properties across data, model capacity, and compute. These works show that learning from kinematic trajectory is increasingly effective for contact-free body motion.

\noindent{\textbf{Manipulation.}}
While locomotion imitation has been increasingly successful, extending physics-based imitation to manipulation remains challenging due to contact-rich dynamics and the difficulty of recovering interaction forces from kinematic data. Early approaches such as InterPhys~\cite{interphys} required task-specific reward design, limiting scalability. PhysHOI~\cite{physhoi} introduced contact graphs for task-agnostic HOI imitation, and SkillMimic~\cite{skillmimic, skillmimicv2} proposed skill-level imitation robust to noisy demonstrations. Scaling further, InterMimic~\cite{intermimic} achieved universal whole-body HOI control through teacher-student distillation with contact-guided rewards, while OmniGrasp~\cite{omnigrasp} leveraged a universal dexterous motion representation to grasp over 1200 objects without paired full-body MoCap data. TokenHSI~\cite{tokenhsi} and CLoSD~\cite{closd} unified diverse interactions through task tokenization and diffusion-physics coupling respectively, and Wu \etal.~\cite{hoifhli} proposed a complete pipeline from language instructions to physically plausible finger-object interactions. At the hand level, Chen \etal.~\cite{chen2024object} propose a hierarchical wrist-guided framework for object-centric dexterous manipulation from human motion data. 
GraspXL~\cite{graspxl} further studies large-scale grasp motion generation by learning policies that produce object-conditioned grasping trajectories across diverse objects and hand configurations. In contrast, we address whole-body HOI retargeting with decoupled supervision in a single policy. DexMachina~\cite{dexmachina}, ManipTrans~\cite{maniptrans}, and SPIDER~\cite{spider} further advance dexterous retargeting through object-state tracking, residual transfer learning, and physics-informed embodiment bridging, respectively. Despite these advances, existing approaches either rely on dense kinematic supervision of detailed finger motion or focus on hand-level manipulation without integrating full-body control.

\subsection{Dexterous Manipulation with Multi-Fingered Hands}
\noindent{\textbf{Grasp Pose Generation.}}
Early learning-based dexterous grasp generation benefited from differentiable contact modeling. Grasp'd~\cite{graspd} synthesizes multi-finger grasps by optimizing through differentiable contact-rich simulation. The availability of large-scale synthetic grasp datasets such as DexGraspNet~\cite{dexgraspnet} shifted the field toward data-driven prediction of stable grasp poses. Subsequent works improved coverage and multimodality with proposal-and-policy formulations~\cite{unidexgrasp} and generative modeling, including diffusion and transformer-based grasp synthesis~\cite{dexdiffuser, ugg, dgtr}. Beyond purely geometric conditioning, recent methods incorporate task intent via language. DexGYS~\cite{dexgys} enables language-guided dexterous grasp generation, while SemGrasp~\cite{semgrasp} and DexVLG~\cite{dexvlg} further align grasp representations with vision-language priors. Dexonomy~\cite{dexonomy} organizes grasps into a taxonomy and AffordDexGrasp~\cite{afforddexgrasp} targets open-set language-guided affordances. DextER~\cite{dexter} introduces contact-based embodied reasoning by predicting which finger links should contact where on the object surface before generating the final grasp, improving intention alignment and enabling steerable grasp synthesis. While these methods generate text and affordance-aligned grasp poses, they focus primarily on the final grasp configuration, rather than learning physics-aware trajectories or accurate manipulation sequences.

\noindent{\textbf{Dexterous Manipulation Learning.}}
Beyond static grasp pose generation, dexterous manipulation learning targets temporally extended, closed-loop control for contact-rich multi-fingered hands. Early successes with large-scale reinforcement learning demonstrated agile in-hand manipulation via massive simulation and domain randomization~\cite{rubiks, andrychowicz2020learning}. Imitation learning offers an alternative supervision source: Qin \etal.~\cite{qin2022one} transfers single-camera teleoperation to multiple hand embodiments, while EgoMimic~\cite{egomimic} scales policy learning from egocentric human videos. Subsequent work improved robustness and horizon length, including DeXtreme~\cite{dextreme} for sim-to-real agile in-hand skills and Sequential Dexterity~\cite{chen2023sequential} for chaining dexterous policies over long-horizon tasks. To better leverage contact information, RotateIt~\cite{rotateit} integrates vision and tactile for general in-hand rotation. Recently, foundation-model-driven policies have begun to couple task-level reasoning with low-level dexterity: DexGraspVLA~\cite{dexgraspvla} combines a VLM-based planner with a diffusion action controller trained by imitation for robust, instruction-following dexterous grasping, and DexUMI~\cite{dexumi} explores using human hands as a universal manipulation interface for scalable demonstration collection. EgoScale~\cite{egoscale} shows that scaling egocentric human data can serve as predictable supervision for dexterous manipulation by pretraining on large-scale wrist and retargeted hand actions and transferring to real robots with minimal robot data. These hand-centric advances motivate our study of how to integrate dexterous manipulation into whole-body retargeting while reducing reliance on finger kinematic references.
\label{sec:relwork}

\section{Method}
\label{sec:method}
\begin{wrapfigure}{r}{0.5\textwidth}
\centering

\scalebox{0.48}{
    \begin{minipage}{\columnwidth}
    \centering
    \includegraphics[width=1.0\columnwidth]{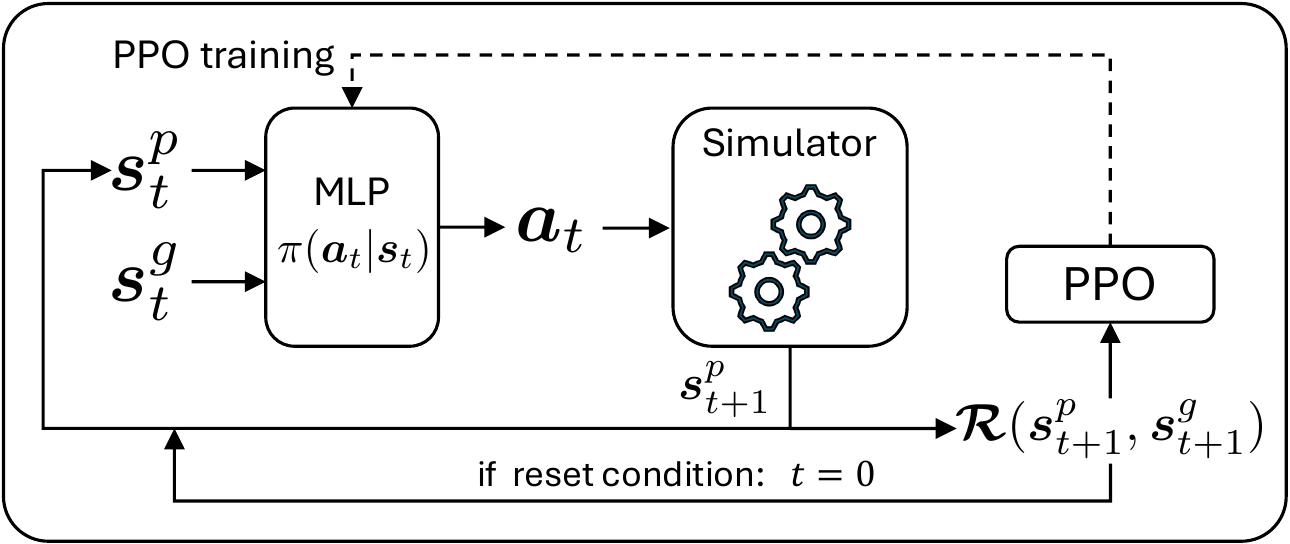}
    \end{minipage}
}

\vspace{-2mm}
\captionsetup[figure]{font=normalsize}
\captionof{figure}{\textbf{Training pipeline.} The policy observes proprioceptive and goal state to generate actions. Rewards measure alignment with reference motion using time-varying weights that prioritize contacts and object dynamics. \Alg{wristmimic} details the complete training procedure.}
\label{fig:overall_architecture}

\vspace{2mm}

\scalebox{0.46}{
    \begin{minipage}{\columnwidth}
    \hrule
    \vspace{2mm}
    \captionsetup[algorithm]{font=LARGE}
    \captionof{algorithm}{WristMimic Training}
    \vspace{-1mm}
    \label{alg:wristmimic}
    \hrule
    \begin{algorithmic}[1]
    \Large
    \State \textbf{Input}: Reference $\{\hat{\mathbf{q}}_t, \hat{\mathbf{q}}^{\text{obj}}_t, \hat{\mathbf{c}}_t\}$, contact frame $t_c$
    \State \textbf{Preprocessing}: Identify grasping hands $\mathcal{J}_{\text{arm}}$
    \For{episode $= 1$ to $N_{\text{episodes}}$}
        \State Sample $\boldsymbol{s}_0$ from reference
        \While{$t < T_{\text{max}}$ and not terminated}
            \State $\boldsymbol{s}_t^p \gets \text{get\_proprioception()}$
            \State $\boldsymbol{s}_t^g \gets \text{get\_goal()}$ (excludes 30 finger joints)
            \State $\boldsymbol{a}_t \sim \pi(\cdot|\boldsymbol{s}_t^p, \boldsymbol{s}_t^g)$ (actions for 51 joints)
            \State $\boldsymbol{s}_{t+1}^p \gets \text{sim.step}(\boldsymbol{a}_t)$
            \State Compute time-varying weights:
            \State \quad $w_j(t) = \begin{cases}
                0 & j \in \mathcal{J}_{\text{arm}},\ t \in [t_c - t_b, t_c + t_a] \\
                1 & j \in \mathcal{J}_{\text{wrist}},\ t \in [t_c - t_b, t_c + t_a] \\
                w_{\text{red}} & j \in \mathcal{J}_{\text{cf}} \setminus
                (\mathcal{J}_{\text{arm}} \cup \mathcal{J}_{\text{wrist}}),\
                \\ &t \in [t_c - t_b, t_c + t_a] \\
                1 & \text{otherwise}
            \end{cases}$
            \State Compute reward components:
            \State \quad $r^\alpha_t = \exp(-\lambda^\alpha \sum_j w_j(t) E_j^\alpha)$ where $\alpha \in \{p, \theta\}$
            \State \quad $r^{\alpha,\text{obj}}_t = \exp(-\lambda^{\alpha,\text{obj}} E^{\alpha,\text{obj}})$ where $\alpha \in \{p, \theta\}$
            \State \quad $r^c_t = \exp(-\lambda^c E^c)$
            \State Compute multiplicative reward:
            \State \quad $r_t = r^p_t \cdot r^\theta_t \cdot r^{p,obj}_t \cdot r^{\theta,obj}_t \cdot r^c_t$
            \If{base termination conditions met}
                \State terminated $\gets$ True
            \EndIf
            \If{$t \in [t_c - t_b, t_c + t_a]$}
                \State phase $\gets$ get\_phase($t, t_c$)
                \If{wrist error $>$ phase-specific threshold}
                    \State terminated $\gets$ True
                \EndIf
            \EndIf
            \State Store ($\boldsymbol{s}_t, \boldsymbol{a}_t, r_t$)
        \EndWhile
        \State Update $\pi$ and $V$ via PPO when buffer full
    \EndFor
    \end{algorithmic}
    \vspace{1mm}
    \hrule
    \end{minipage}
}

\vspace{-15mm}
\end{wrapfigure}

\noindent{\textbf{Overview.}}
We train a policy network that outputs actions in a physics-based simulator to reproduce a reference human-object interaction sequence. The reproduced motion must \textbf{(i)} track the reference body trajectory and \textbf{(ii)} manipulate the object so that its motion follows the reference kinematic trajectory. For the first one, standard imitation learning is often sufficient as the objective primarily involves joint movement. However, accurate object manipulation cannot be achieved by body pose imitation alone. Motion capture is imperfect, and position-based trajectories do not encode interaction forces; for example, grasping an object with weak or strong force may correspond to identical finger positions. As a result, correct pose imitation does not necessarily imply correct manipulation. Since most manipulation-related contacts occur at the hands, we decouple supervision by contact regime: the policy tracks kinematic pose targets only for object-contact-free joints, including the body and wrists, while finger joints receive no kinematic targets and are instead learned through object motion and contact outcomes, which reflect successful physical interaction. To make this decoupled, object-driven learning formulation reliable, we treat the wrist as the bridge between contact-free body control and contact-rich manipulation. We constrain wrist behavior around contact using a contact-window with time-varying reward weights and phase-specific reset thresholds. Our pipeline is illustrated in \Fig{overall_architecture} and detailed in \Alg{wristmimic}.

\subsection{Problem Formulation.}
Given reference motion data $\{\hat{\mathbf{q}}_t, \hat{\mathbf{q}}^{\text{obj}}_t, \hat{\mathbf{c}}_t\}_{t=1}^T$ including body poses $\hat{\mathbf{q}}_t$, object trajectories $\hat{\mathbf{q}}^{\text{obj}}_t$, and contact labels $\hat{\mathbf{c}}_t$, we learn a physics-based policy $\pi$ which outputs action $\boldsymbol{a}_t$ to generate simulated states $\{\mathbf{q}_t, \mathbf{q}^{\text{obj}}_t, \mathbf{c}_t\}$. 
We formulate this as a MDP $\mathcal{M}=\langle\boldsymbol{S}, \boldsymbol{A}, \boldsymbol{T}, \boldsymbol{\mathcal{R}}, \gamma \rangle$ with state space $\boldsymbol{S}$, action space $\boldsymbol{A}$, simulator transition dynamics $\boldsymbol{T}$, reward $\boldsymbol{\mathcal{R}}$, and discount factor $\gamma$.
The state $\boldsymbol{s}_t = (\boldsymbol{s}_t^p, \boldsymbol{s}_t^g)$ consists of a proprioceptive state $\boldsymbol{s}_t^p$ from simulator and a goal state $\boldsymbol{s}_t^g$ from the reference. The policy $\pi(\boldsymbol{a}_t | \boldsymbol{s}_t)$ predicts actions for all 51 joints, with actions sampled from Gaussian distributions $\mathcal{N}(\boldsymbol{\mu}_t, \sigma^2 \mathbf{I})$. Here $\boldsymbol{\mu}_t \in \mathbb{R}^{51 \times 3}$ are predicted means and $\sigma$ is fixed standard deviation. 
We optimize $\pi$ with PPO~\cite{ppo} to maximize reward $r_t = \boldsymbol{\mathcal{R}}(\boldsymbol{s}_t^p, \boldsymbol{s}_t^g)$ measuring alignment between the simulated state and the reference. Detailed policy representation is in \S\ref{subsec:policy_representation}. 

\subsection{Policy Representation for Decoupled Manipulation}
\label{subsec:policy_representation}
\noindent\textbf{Proprioceptive State.}
The humanoid follows the SMPL-X~\cite{smplx} configuration with 52 rigid bodies connected by 51 actuated joints, each with 3 DoFs. The 51 actuated joints are partitioned into two sets: $\mathcal{J}_{\text{cf}}$ (object-contact-free joints) consisting of 19 body joints and 2 wrists, and $\mathcal{J}_{\text{cr}}$ (contact-rich joints) consisting of 30 finger joints. We denote the root pose as $\mathbf{q}^{\text{root}}_t = (\mathbf{p}^{\text{root}}_t, \boldsymbol{\theta}^{\text{root}}_t) \in \mathbb{R}^{3} \times \mathbb{R}^{6}$ and velocity as $\dot{\mathbf{q}}^{\text{root}}_t = (\mathbf{v}^{\text{root}}_t, \boldsymbol{\omega}^{\text{root}}_t) \in \mathbb{R}^{3} \times \mathbb{R}^{3}$, where $\mathbf{p}$, $\boldsymbol{\theta}$, $\mathbf{v}$, and $\boldsymbol{\omega}$ represent position, orientation, linear velocity, and angular velocity, respectively.
The 51 actuated joints have poses $\mathbf{q}_t \in \mathbb{R}^{51 \times 3} \times \mathbb{R}^{51 \times 6}$ and velocities $\dot{\mathbf{q}}_t \in \mathbb{R}^{51 \times 3} \times \mathbb{R}^{51 \times 3}$ with the same representation. The 52 rigid bodies are used to compute body-level contacts and interaction graph features.
Object state is represented by $\mathbf{q}^{\text{obj}}_t$ with velocities $\dot{\mathbf{q}}^{\text{obj}}_t$.
Binary contact markers $\mathbf{c}_t \in \mathbb{R}^{52}$ indicate contact for each rigid body, and interaction graph vectors $\mathbf{i}_t \in \mathbb{R}^{52\times 3}$ encode the spatial relationship between each humanoid rigid body and its nearest object surface point~\cite{tactile,3dvitac,difftactile,mimictouch}.
The proprioceptive state is:
\begin{equation}
\boldsymbol{s}_t^p = \{\mathbf{q}^{\text{root}}_t, \dot{\mathbf{q}}^{\text{root}}_t, \mathbf{q}_t, \dot{\mathbf{q}}_t, \mathbf{q}^{\text{obj}}_t, \dot{\mathbf{q}}^{\text{obj}}_t, \mathbf{c}_t, \mathbf{i}_t\}.
\end{equation}

\noindent\textbf{Goal State.}
The goal state $\boldsymbol{s}_t^g = \{g_{t,t+k}\}_{k \in K}$ encodes differences between the current state and future reference states at time horizons $K$:
\begin{equation}
g_{t,t+k} = \{\Delta\boldsymbol{\theta}_{t+k}, \Delta\mathbf{p}_{t+k}, \Delta\boldsymbol{\theta}^{\text{obj}}_{t+k}, \Delta\mathbf{p}^{\text{obj}}_{t+k}, \Delta\mathbf{i}_{t+k}, \Delta\mathbf{c}_{t+k}\},
\end{equation}
where $\Delta$ denotes the difference between the reference state at $k$ frames ahead and the current state, using $\ominus$ for rotation differences and $-$ for others. For decoupled supervision, we assign kinematic pose targets only to joints in $\mathcal{J}_{\text{cf}}$ and exclude all joints in $\mathcal{J}_{\text{cr}}$. No kinematic reference trajectories are provided for finger joints. As a result, finger behavior is not supervised through pose imitation, but is instead optimized indirectly through object motion and contact objectives, \ie, via the object pose targets 
$\Delta\boldsymbol{\theta}^{\text{obj}}_{t+k}$ and $\Delta\mathbf{p}^{\text{obj}}_{t+k}$, together with contact alignment rewards. The contact target $\Delta\mathbf{c}_{t+k} \in \mathbb{R}^{23}$ consists of 21 explicit contact labels for the non-finger rigid bodies and 2 binary hand-level contact indicators (one per hand). A hand-level contact indicator is set to 1 if any rigid body belonging to that hand is in contact with the object. All goal elements are normalized relative to the root position and facing direction~\cite{omnigrasp, pulse, intermimic, skillmimic, skillmimicv2, maskedmanipulator,PMP,phc}.

\noindent\textbf{Reward.}
Our reward function follows a multiplicative structure:
\begin{equation}
r_t = r^p_t \cdot r^{\theta}_t \cdot r^{p,\text{obj}}_t \cdot r^{\theta,\text{obj}}_t \cdot r^c_t,
\end{equation}
where each component takes the form $r = \exp(-\lambda E)$, penalizing error $E$ between simulated and reference states with component-specific weights $\lambda$. The five components measure: joint position tracking ($r^p_t$), joint rotation tracking ($r^{\theta}_t$), object position tracking ($r^{p,\text{obj}}_t$), object rotation tracking ($r^{\theta,\text{obj}}_t$), and contact alignment ($r^c_t$). 
Following InterMimic~\cite{intermimic}, $E^p$ and $E^\theta$ prioritize rotation when joints are far from the object and position when joints are near.

To implement decoupled supervision, we compute pose-tracking errors ($r^p_t$, $r^{\theta}_t$) only over joints in $\mathcal{J}_{\text{cf}}$, and exclude all 30 finger joints from direct kinematic tracking. For contact alignment, we compute error over the 23 contact elements defined in the goal state (21 non-finger bodies + 2 hand-level indicators). 
As a result, contact-free body parts receive reward signals from kinematic references, while finger joints are influenced by object pose tracking and contact alignment. This reward structure aligns with our decoupled formulation, where finger joints are not directly supervised by pose targets but are shaped by object motion and contact alignment.

\noindent\textbf{Action.}
Actions $\boldsymbol{a}_t \in \mathbb{R}^{51 \times 3}$ specify target rotations for all 51 actuated joints in exponential map coordinates~\cite{intermimic}, which are converted to torques via PD controllers.
Although finger joints receive no kinematic supervision in the goal state, they remain fully actuated in the action space. Their behavior is learned implicitly through object pose rewards, contact alignment, and interaction dynamics.

\subsection{Learning Wrist-Guided Manipulation}
\label{subsec:wrist_manipulation}
After decoupling kinematic supervision for contact-free body parts from object- and contact-guided finger behavior, we identify the wrist as the structural bridge between these two regimes. While finger behavior is learned through object motion and contact outcomes, effective exploration requires the hand to be placed in an affordance-consistent and reachable configuration. Since the wrist determines the global pose of the fingers and the reachable contact set, a misaligned wrist may prevent access to the intended contact region. Moreover, as the wrist is typically free from direct object contact, it remains suitable for kinematic guidance using reference trajectories. We therefore prioritize wrist placement through selective relaxation of proximal arm supervision and wrist-specific phase-dependent reset thresholds. Finger configurations are then allowed to emerge from interaction dynamics.

\noindent\textbf{Prioritized Alignment Strategy.}
To ensure feasible wrist placement that enables finger behavior to emerge through interaction dynamics, we adopt a prioritized alignment strategy. Motion capture data is inherently imperfect when executed in physics-based simulation, making it difficult to simultaneously align all body joints with the reference motion. For example, motion capture sequences are often defined with a floating root frame that may violate physical constraints in simulation. As a result, some joints must be allowed to deviate to achieve accurate wrist alignment. We define a \textit{contact window} $[t_c - t_b, t_c + t_a]$ centered on the first contact frame $t_c$, where $t_b$ and $t_a$ denote frames before and after contact. Within this window, we enforce tight wrist constraints while relaxing upper arm supervision, allowing the policy to adjust upper arm configurations freely to position the wrist optimally. This strategy is implemented through two complementary mechanisms: reward weight modulation and phase-specific reset boundaries. 

\noindent\textbf{Reward Weight Modulation.}
To prioritize wrist alignment while relaxing upper arm supervision, we modulate joint reward weights over time. Specifically, the joint rewards are formulated to exclude all 30 finger joints:
\begin{equation}
r^\alpha_t = \exp\left(-\lambda^\alpha \sum_{j \in \mathcal{J}_{\text{cf}}} w_j(t) E_j^\alpha\right) \quad \text{where } \alpha \in \{p, \theta\},
\end{equation}
where $\mathcal{J}_{\text{cf}}$ denotes the 19 body joints and 2 wrist joints, and $w_j(t)$ are time-varying weight factors defined as:
\begin{equation}
w_j(t) =
\begin{cases}
0, & \text{if } j \in \mathcal{J}_{\text{arm}},\ t \in [t_c - t_b, t_c + t_a], \\
1, & \text{if } j \in \mathcal{J}_{\text{wrist}},\ t \in [t_c - t_b, t_c + t_a], \\
w_{\text{red}}, & \text{if } j \in \mathcal{J}_{\text{cf}} \setminus
(\mathcal{J}_{\text{arm}} \cup \mathcal{J}_{\text{wrist}}),\
t \in [t_c - t_b, t_c + t_a], \\
1, & \text{otherwise}.
\end{cases}
\label{eq:reward_weight_modulation}
\end{equation}
where $\mathcal{J}_{\text{arm}}$ denotes the shoulder and elbow joints of the grasping hand(s), $\mathcal{J}_{\text{wrist}}$ denotes the wrist joints, and $w_{\text{red}} < 1$ is the reduced weight factor. This selective relaxation removes proximal arm constraints ($w_j=0$), keeps wrist supervision unchanged ($w_j=1$), and reduces the weights of the remaining contact-free body joints ($w_j=w_{\text{red}}$) during the contact window. Moreover, reducing other body joint weights ($w_j=w_{\text{red}}$) allows the policy to focus on hand-object contact and object dynamics in contact window. Specific weight values are detailed in the supplementary material.
\captionsetup[figure]{skip=2pt}
\begin{figure*}[h]
    \centering
    \includegraphics[width=\textwidth]{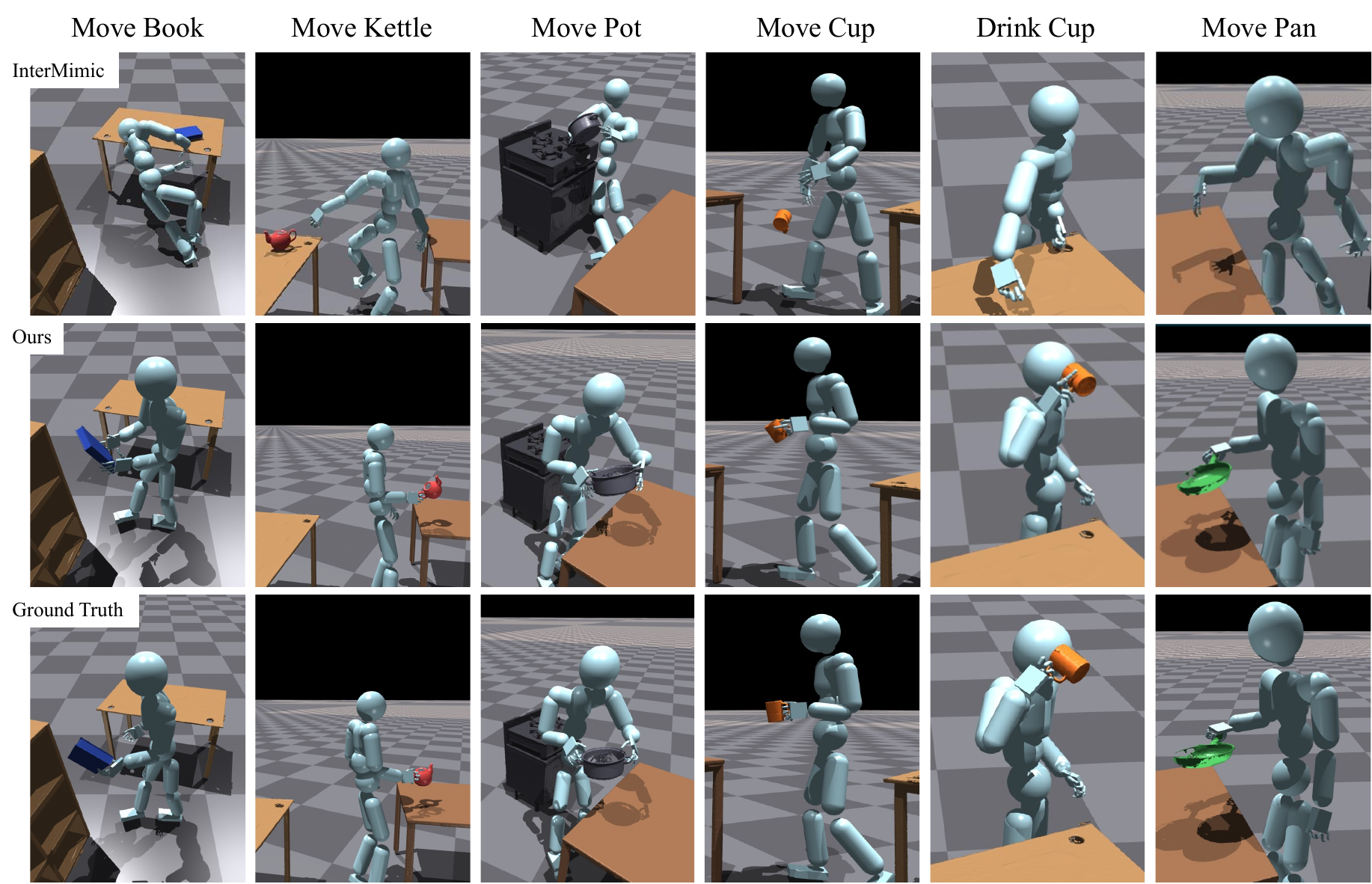}  
     \caption{\textbf{Qualitative comparison across 6 representative scenes.} We illustrate representative qualitative results comparing our method with InterMimic~\cite{intermimic}. Our method achieves affordance-aware grasping with diverse manipulation behaviors while accurately reproducing whole-body motion in the physics-based simulation.}
    \label{fig:qualitative_results}
    \vspace{-5mm}
\end{figure*}

\noindent\textbf{Phase-Specific Reset Thresholds.}
We employ early termination to prevent inefficient exploration when episodes diverge from feasible trajectories. Specifically, episodes terminate if: \textbf{(i)} body pose deviates severely from reference, \textbf{(ii)} root height drops below a threshold (indicating fall), \textbf{(iii)} object pose drifts too far from reference, or \textbf{(iv)} contact error persists for multiple frames~\cite{intermimic,omnigrasp}. In addition to this base reset strategy, we apply phase-specific reset thresholds to the wrist of grasping hand(s) within the contact window. We subdivide the contact window into three phases centered on $t_c$: \textit{approach} $[t_c - t_b, t_c + \tau_1]$, \textit{grasping} $[t_c + \tau_1, t_c + \tau_2]$, and \textit{stabilization} $[t_c + \tau_2, t_c + t_a]$, where $0 \leq \tau_1 < \tau_2 \leq t_a$ define phase transitions after first contact. During the \textit{grasping} phase, we apply tight thresholds of 7cm in position and 0.2 radians in orientation, terminating if wrist deviates beyond these bounds from reference (\Fig{teaser}, top right). The \textit{approach} and \textit{stabilization} phases use looser thresholds of 15cm and 0.5 radians, allowing trajectory exploration while maintaining approximate alignment. This design—using reduced reward weights with strict reset boundaries rather than high reward weights—enables the policy to discover optimal wrist poses within a feasible search space, accommodating mocap inaccuracies while ensuring natural grasp configurations.

\begin{table}[t]
\centering
\scriptsize
\begin{tabularx}{0.95\textwidth}{l *{3}{>{\centering\arraybackslash}X}}
\toprule
& \multicolumn{3}{c}{Success rate (\%) $\uparrow$ / Obj. pos. err. (cm) $\downarrow$ / Obj. rot. err. ($^{\circ}$) $\downarrow$} \\
\cmidrule(lr){2-4}
\textbf{Method} & OMOMO~\cite{omomo} & ParaHome~\cite{parahome} & Avg. \\
\midrule
InterMimic~\cite{intermimic} &  86.6 / 14.2 / 22.2 & 0.1 / 83.5 / 73.3 & 43.3 / 48.8 / 47.7 \\ 
SkillMimicV2~\cite{skillmimicv2} & - & 1.3 / 72.1 / 80.3 & 1.3 / 72.1 / 80.3 \\ 
WristMimic (Ours)                  & \textbf{98.9} / \textbf{7.3} / \textbf{12.2} & \textbf{83.3} / \textbf{15.3} / \textbf{33.9} & \textbf{91.1} / \textbf{11.3} / \textbf{23.1} \\ 
\bottomrule
\end{tabularx}
\vspace{2mm}
\caption{\textbf{Comparison with state-of-the-art methods.} We compare our method against recent WBC-based object interaction baselines. Results are reported across 40 diverse interaction sequences sampled from OMOMO (20)~\cite{omomo}, and ParaHome (20)~\cite{parahome}. Note that SkillMimicV2~\cite{skillmimicv2} is evaluated only on ParaHome, as it specifically utilizes bone-vector representations. \textbf{Bold} indicates the best performance.}
\label{tab:sota_comparison}
\vspace{-5mm}
\end{table}

\section{Experiments}
\label{sec:experiments}
\subsection{Experimental Setup}
\noindent\textbf{Datasets and Evaluation Metrics.}
We evaluate WristMimic on two human–object interaction datasets: ParaHome~\cite{parahome} and OMOMO~\cite{omomo}. These datasets provide object and subject geometries, human motion trajectories represented via SMPL-X~\cite{smplx} or bone vectors, and corresponding object trajectories. Each episode spans approximately 100–250 frames. We train per-scene policies on 20 representative sequences from ParaHome and OMOMO respectively. This setup allows us to evaluate the robustness of our wrist-guided, object-kinematics-driven approach across diverse manipulation scenarios. We assess performance using three primary metrics across 10,000 rollout trials per scene:
\begin{itemize}
    \item Success rate (\%): A trial is successful if the sequence reaches completion without early termination, the average object position error remains below 10 cm, and the robot maintains contact for at least 80\% of the reference contact duration.
    \item Object position/rotation error (cm, $\degree$): Aligning with our principle that manipulation imitation should prioritize object trajectory over exact finger replay, we measure the average position/rotation deviation of the object from its reference trajectory throughout the sequence.
\end{itemize}
 
\noindent\textbf{Implementation Details.}
We implement WristMimic in Isaac Gym~\cite{gym} and train all policies using Proximal Policy Optimization (PPO)~\cite{ppo}. Training uses 2048 parallel environments on a single NVIDIA RTX 3090 GPU with 30 Hz control frequency. The policy network consists of separate actor and critic MLPs with SiLU~\cite{silu} activations. Contact window parameters are set to $t_b = 10$, $t_a = 15$ frames, with phase transitions at $\tau_1 = -2$ and $\tau_2 = 12$ frames relative to the first contact frame $t_c$. Complete hyperparameters are provided in the supplementary material.   

\subsection{Quantitative Results}
\label{subsec:quantitative_results}
We compare WristMimic against InterMimic~\cite{intermimic} and SkillMimicV2~\cite{skillmimicv2}. For fair comparison, we train per-scene policies for all methods. Evaluation is conducted on 40 sequences in total, consisting of 20 sequences from ParaHome~\cite{parahome} and 20 from OMOMO~\cite{omomo}. For SkillMimicV2, evaluation is performed only on ParaHome, since SkillMimicV2 requires bone-vector representations, which are available only in ParaHome.

Quantitative results are shown in \Tbl{sota_comparison}. WristMimic achieves the best performance across all evaluation metrics despite not relying on explicit finger pose supervision. In particular, on the OMOMO dataset~\cite{omomo}, WristMimic reduces both object position and rotation errors by nearly half compared to recent methods. These results indicate that guiding manipulation through wrist placement and object kinematics provides a strong structural signal for object-trajectory–aware control.

On the ParaHome dataset~\cite{parahome}, both InterMimic and SkillMimicV2 exhibit significantly lower success rates. Notably, while InterMimic shows a clear performance gap between OMOMO and ParaHome, WristMimic maintains relatively consistent performance across the two datasets. Since ParaHome contains more dexterous interaction scenarios such as grasping small handles or objects with narrow contact regions, these results suggest that wrist-guided constraints provide robust structural guidance for delicate grasping.

\captionsetup[figure]{skip=2pt}
\begin{figure*}[h]
    \centering
    \includegraphics[width=\textwidth]{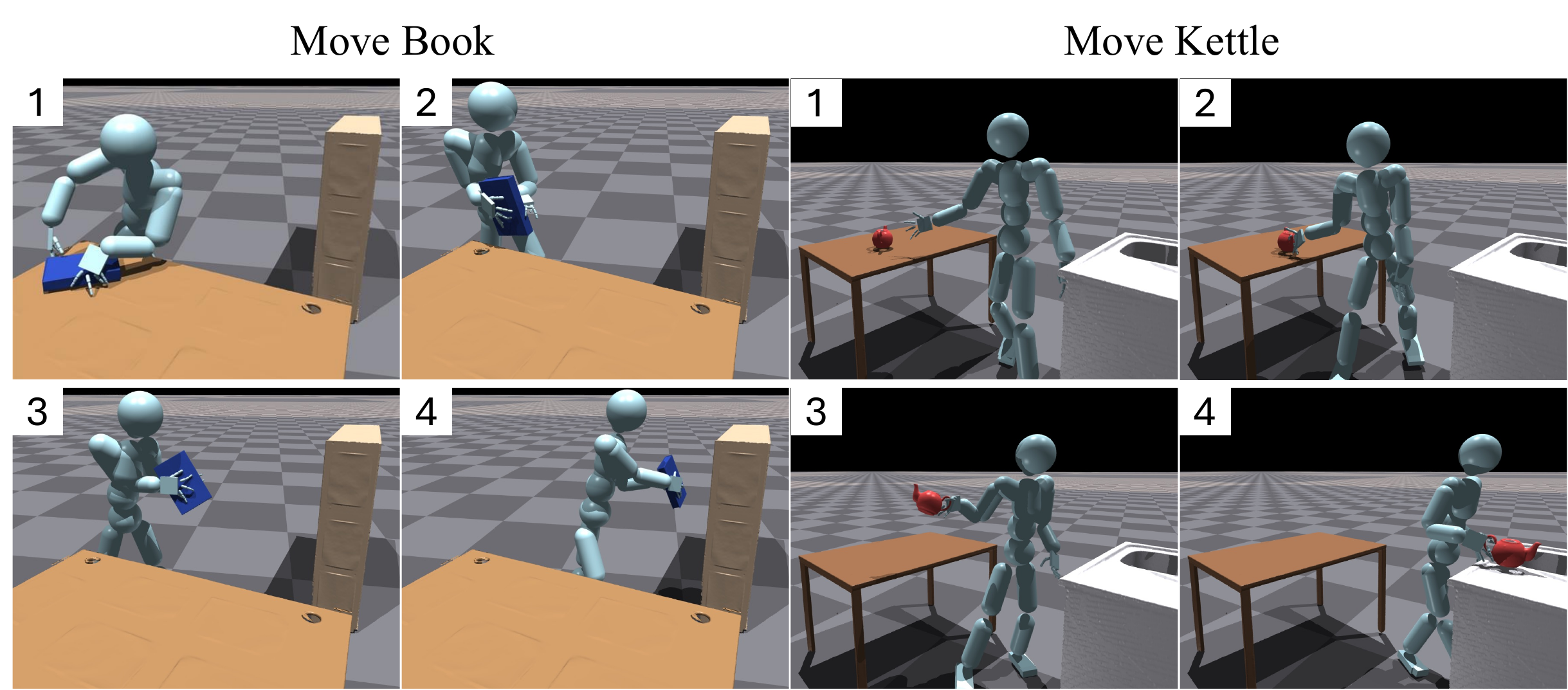}
    \vspace{1mm}
    \caption{\textbf{Key frames from manipulation sequences.} Our method successfully handles challenging manipulation scenarios, including grasping a book from a flat table without a handle and grasping a kettle followed by turning the body while maintaining a stable grasp.}
  \label{fig:seq_qual}
    \vspace{-10mm}
\end{figure*}

\subsection{Qualitative Results}
\begin{wrapfigure}{r}{0.6\textwidth} 
\vspace{-10pt}
\centering

\scalebox{0.6}{
    \begin{minipage}{\columnwidth}
    \centering
    \includegraphics[width=\columnwidth]{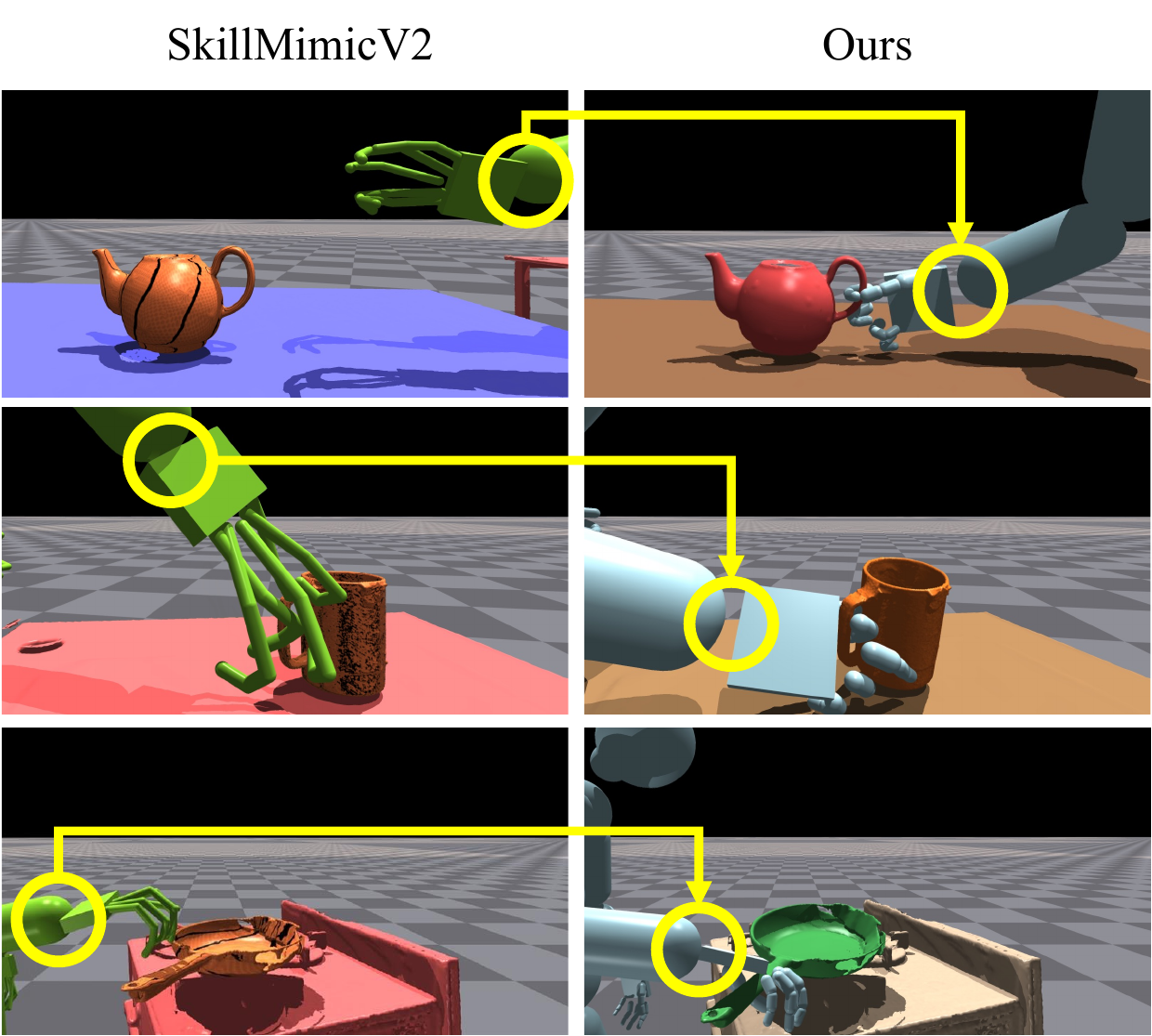}
    \end{minipage}
}

\captionsetup[figure]{font=normalsize}
\captionof{figure}{\textbf{Grasping pose comparison.} Although SkillMimicV2 demonstrates plausible finger closure, the incorrect wrist pose prevents a successful grasp, whereas WristMimic maintains precise wrist alignment.}
\label{fig:wrong_wrist_correct_finger}

\vspace{-20pt}
\end{wrapfigure}

We present representative qualitative results comparing our method with InterMimic~\cite{intermimic} on six scenes from the ParaHome dataset (\Fig{qualitative_results}). As discussed in \S\ref{subsec:quantitative_results}, previous methods often struggle in these dexterous manipulation scenarios. In particular, as illustrated in \Fig{wrong_wrist_correct_finger}, even when SkillMimicV2 produces visually plausible finger configurations, incorrect wrist placement prevents stable object grasping. This example highlights that correct wrist placement is a prerequisite for establishing feasible contact configurations for grasping. In contrast, our approach first establishes a feasible wrist configuration, allowing finger poses to naturally adapt to the object and enabling both stable grasping and accurate tracking of the reference object trajectory. These examples demonstrate that guiding manipulation through wrist placement enables affordance-consistent grasping in dexterous interaction scenarios.

\noindent\Fig{seq_qual} further shows sequential frames sampled at 50-frame intervals from two representative manipulation sequences. WristMimic maintains stable object contact while producing natural hand poses throughout the sequence, demonstrating consistent manipulation behavior over time.

In the Move Book scenes in \Fig{qualitative_results} and \Fig{seq_qual}, a slight deviation from the ground truth motion can be observed. The human demonstration initially lifts the book using both hands before releasing one hand, whereas our policy continues to support the book with both hands due to grasping stability. This behavior reflects our method’s design choice to prioritize stable object manipulation and trajectory tracking over strict imitation of the human body pose.

\begin{table*}[t]
\centering
\resizebox{0.95\textwidth}{!}{
\begin{tabular}{lc} 
\toprule
& \multicolumn{1}{c}{Success rate(\%) $\uparrow$ / Obj. pos. err.(cm) $\downarrow$} / Obj. rot. err.($\degree$) $\downarrow$ \\
\cmidrule(lr){2-2}
\textbf{Method} & ParaHome~\cite{parahome} \\
\midrule
Decoupled (No Wrist Constraints) & 0.0 / 56.4 / 49.7 \\
+ reward weight modulation only  & 0.0 / 50.3 / 43.6 \\
+ phase specific reset only      & 67.5 / 28.8 / 56.9 \\
WristMimic (Ours)                & \textbf{86.5} / \textbf{9.9} / \textbf{36.0} \\
\bottomrule
\end{tabular}
}
\vspace{2mm}
\caption{\textbf{Ablation on wrist constraints.}
We evaluate the effect of removing wrist-specific constraints on the ParaHome dataset. WristMimic significantly improves both success rate and object pose accuracy compared to the decoupled formulation without wrist constraints, highlighting the importance of wrist guidance for stable object manipulation. \textbf{Bold} indicates the best performance.}
\label{tab:wrist_ablation}
\vspace{-5mm}
\end{table*}

\subsection{Ablation Study}

\noindent\textbf{Ablation on Wrist Constraints.}
We evaluate the effect of wrist-specific constraints on 8 representative ParaHome~\cite{parahome} sequences. As shown in \Tbl{wrist_ablation}, the decoupled formulation without wrist constraints performs significantly worse, indicating that decoupling alone is not sufficient for reliable manipulation. We further decompose our wrist-centric design into two components: reward weight modulation within the contact window and phase-specific wrist reset thresholds. Phase-specific resets provide the main improvement by keeping exploration near feasible wrist poses, while reward weight modulation alone gives only marginal gains. However, combining reward modulation with phase-specific resets yields the best overall performance, substantially improving success rate and object pose accuracy. These results show that WristMimic’s gains come from explicitly guiding wrist placement during contact-rich phases, which is critical for stabilizing hand--object interactions and achieving affordance-consistent grasps.

\noindent\textbf{Ablation on Reset Threshold.}
We further ablate the grasping-phase wrist reset thresholds on four representative ParaHome~\cite{parahome} scenes. As shown in \Tbl{reset_thr_ablation}, 3.5 cm/0.1 rad (tight) is too restrictive for exploration and pre-grasp recovery, while 15 cm/0.5 rad (loose) admits implausible wrist states that destabilize contact. Our default 7 cm/0.2 rad achieves the best trade-off between feasible exploration and reliable wrist placement.

\begin{figure}[t]
\centering
\begin{minipage}[t]{0.40\linewidth}
    \centering
    \setlength{\tabcolsep}{1pt}
    \begin{tabular}{cc}
        \includegraphics[width=0.47\linewidth]{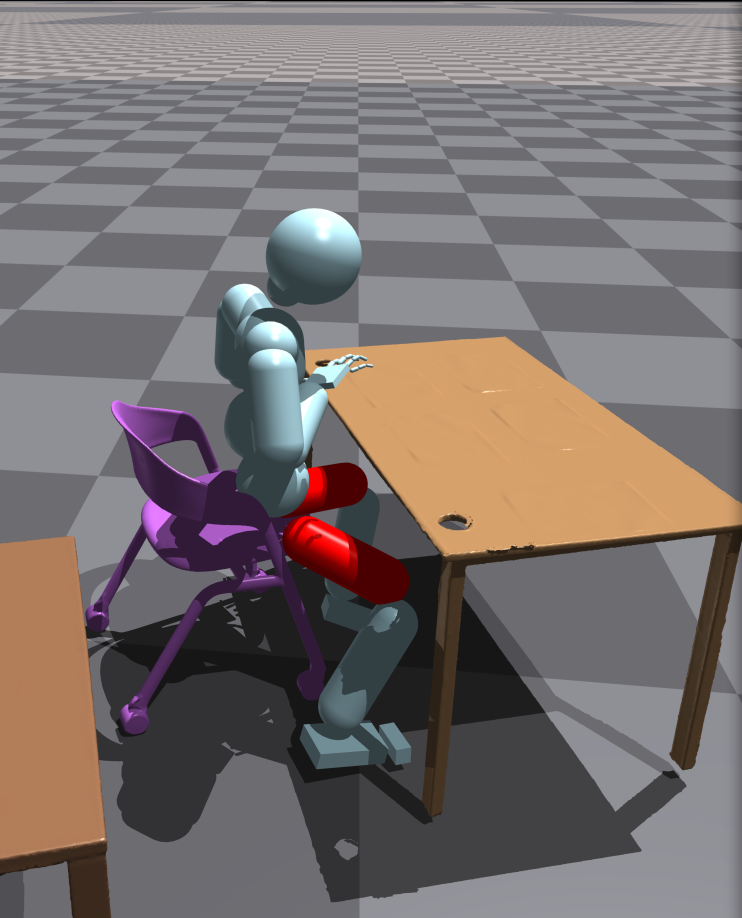}
        &
        \includegraphics[width=0.50\linewidth]{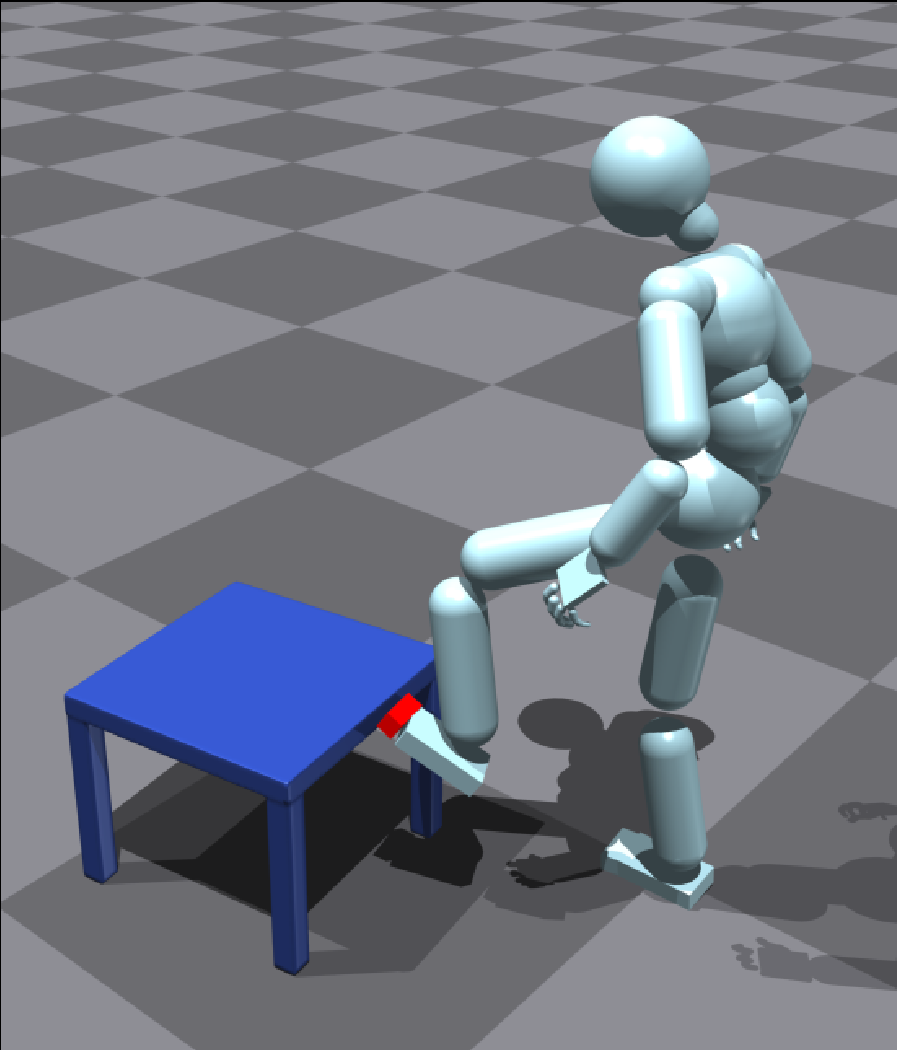}
    \end{tabular}
    \vspace{0.5mm}
    \refstepcounter{figure}\label{fig:non_hand_body_contact}%
    \vspace{-1mm}
    {\footnotesize \parbox[t]{\linewidth}{\textbf{Figure~\thefigure.} Scenes involving non-hand body contact.}}
    \vspace{-2mm}
\end{minipage}
\hfill
\begin{minipage}[t]{0.58\linewidth}
    \centering
    \scriptsize
    \begingroup
    \setlength{\tabcolsep}{2.5pt}
    \resizebox{\linewidth}{!}{%
    \begin{tabular}{lccc}
        \toprule
        Scene & SR (\%) $\uparrow$ & Pos. err. (cm) $\downarrow$ & Rot. err. ($^\circ$) $\downarrow$ \\
        \midrule
        Largetable & 99.9 & 9.9 & 10.6 \\
        Whitechair & 99.7 & 8.0 & 6.8 \\
        Sit on chair \#1 & 97.2 & 9.8 & 5.0 \\
        Sit on chair \#2 & 99.8 & 8.5 & 6.1 \\
        \bottomrule
    \end{tabular}%
    }
    \endgroup
    \par\vspace{0.5mm}
    \refstepcounter{table}
    \label{tab:non_hand_body_contact}
    {\footnotesize \parbox[t]{\linewidth}{\textbf{Table~\thetable.} Rows 1--2: pushing object with foot. Rows 3--4: sit on chair.}}
\end{minipage}
\vspace{-6mm}
\end{figure}

\begin{wrapfigure}{r}{0.5\textwidth}
    \centering
    \vspace{-6mm}
    \begingroup
    \setlength{\tabcolsep}{3pt}
    \renewcommand{\arraystretch}{1.05}
    \resizebox{0.98\linewidth}{!}{%
    \begin{tabular}{lccc}
        \toprule
        Variant & SR (\%) $\uparrow$ & Pos. err. (cm) $\downarrow$ & Rot. err. ($^\circ$) $\downarrow$ \\
        \midrule
        (a) Tight          & 0.0   & 99.8 & 86.8 \\
        (b) \textbf{Ours}  & \textbf{96.3} & \textbf{9.5} & \textbf{18.1} \\
        (c) Loose          & 17.5  & 78.3 & 72.3 \\
        \bottomrule
    \end{tabular}%
    }
    \endgroup
    \vspace{1mm}
    \refstepcounter{table}
    {\small
    \textbf{Table~\thetable.}
    \textbf{Reset threshold ablation.}
    }
    \label{tab:reset_thr_ablation}
    \vspace{-8mm}
\end{wrapfigure}
\noindent\textbf{Ablation on sequences of non-hand-contact bodies.}
\Fig{non_hand_body_contact} and \Tbl{non_hand_body_contact} present ablation studies on non-hand body contact scenarios, where body parts other than the hands serve as the primary contact regions. Unlike hand-object manipulation, these interactions mainly involve coarse body-level support or displacement, and thus can be handled by the standard whole-body mimicking objective without additional task-specific design. The results show that such non-hand contacts do not require any wrist-like design, allowing our framework to extend naturally to scenarios such as sitting on a chair and displacing a small table with the foot.

\noindent\textbf{Hand Morphology Generalization.}
\Tbl{ablation_hand_morphology} demonstrates that our finger-agnostic formulation enables learning across different hand morphologies without requiring finger-specific motion data. We retrain policies using three hand configurations: the default scene-specific SMPL-X hand, the hand model from InterMimic~\cite{intermimic}, and the OmniGrasp~\cite{omnigrasp} hand, evaluated on 10 sequences from ParaHome~\cite{parahome}. These configurations differ substantially in hand size, joint lengths, and initial wrist and finger poses. Despite these differences, grasping remains successful across all configurations, suggesting that wrist guidance provides a strong structural signal for affordance-aware grasping.

\begin{figure*}[t!]
    \centering
    \includegraphics[width=0.8\textwidth]{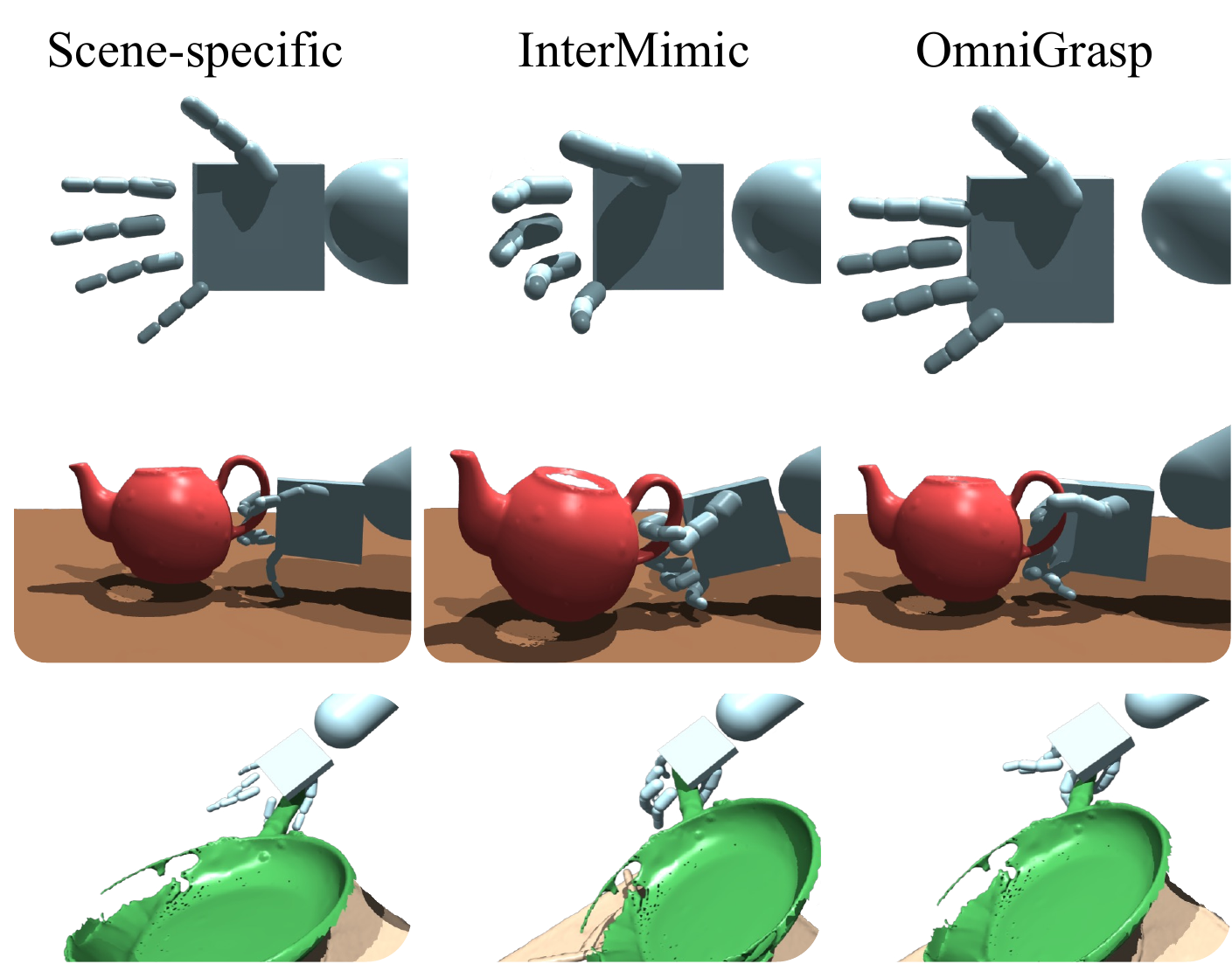}
    \vspace{1mm}
      \caption{\textbf{Ablation on hand-generalization.} Our finger-agnostic approach generalizes across different hand morphologies without finger-specific data. We demonstrate successful grasping with three different hands that differ in joint length, size, and pose.}
  \label{fig:hand_generalization}
    \vspace{-5mm}
\end{figure*}

\Fig{hand_generalization} visualizes representative successful manipulations with the three hand types. All policies are trained independently using the same wrist-guided strategy without explicit finger pose supervision. As shown in \Tbl{ablation_hand_morphology}, performance remains comparable across different hand morphologies, indicating that the policy adapts to varying hand geometries through interaction dynamics rather than kinematic supervision. Among the three configurations, the InterMimic~\cite{intermimic} hand exhibits the most realistic finger poses and highest performance due to its constrained joint ranges that prevent physically implausible poses while other hands set joint range as $[-\pi, \pi]$. Since the joint range-constrained setting is more realistic in the real world, this shows that the wrist-only approach can make natural grasping poses emerge while maintaining grasping quality.
\begin{table*}[t]
\centering
\resizebox{0.95\textwidth}{!}{
\begin{tabular}{lc} 
\toprule
& \multicolumn{1}{c}{Success rate(\%) $\uparrow$ / Obj. pos. err.(cm) $\downarrow$} / Obj. rot. err.($\degree$) $\downarrow$ \\
\cmidrule(lr){2-2}
\textbf{Method} & ParaHome~\cite{parahome} \\
\midrule
InterMimic~\cite{intermimic} & 95.2 / 10.8 / 28.1 \\
OmniGrasp~\cite{omnigrasp}  & 76.4 / 16.9 / 27.1\\
Scene-specific       & 75.8 / 20.1 / 39.6 \\
\bottomrule
\end{tabular}
}
\vspace{2mm}
\caption{\textbf{Hand morphology generalization performance.} 
Scene-specific denotes the default hand following the scene-specific SMPL-X~\cite{smplx} configuration. We evaluate the same policy across different hand morphologies, including the InterMimic and OmniGrasp hand models, which vary in hand size, joint length, and joint limits. }
\label{tab:ablation_hand_morphology}
\vspace{-6mm}

\end{table*}
\captionsetup[figure]{skip=2pt}
\subsection{Limitations}
Our method focuses on affordance-aware grasping and object manipulation rather than precise in-hand manipulation that requires detailed finger pose control. While wrist-guided learning enables the policy to discover natural grasp configurations through interaction dynamics, tasks involving fine-grained finger repositioning or object reorientation within the hand remain beyond the current scope. Extending the framework to support such in-hand manipulation without explicit finger supervision is an interesting direction for future work. 

Another limitation is that our current framework trains scene-specific policies, requiring separate training for each scene. This contrasts with methods such as OmniGrasp~\cite{omnigrasp}, InterMimic~\cite{intermimic}, and MaskedManipulator~\cite{maskedmanipulator}, which aim to generalize across scenes. Nevertheless, our experiments show that the wrist-guided formulation generalizes across different hand morphologies within the same scene, suggesting that the learned interaction structure is not tied to a specific hand model. Developing scene-general policies that retain this wrist-centric interaction structure remains an important direction for future research.

\section{Conclusion}
\label{sec:conclusion}
We have presented WristMimic, a wrist-guided framework for physics-based retargeting of human–object interaction demonstrations. Our approach decouples supervision between contact-free body motion and contact-rich hand manipulation, and highlights the wrist as a key bridge between these two regimes. Because the wrist largely remains contact-free while determining the global configuration of the fingers, guiding the wrist provides a useful structural signal for learning manipulation behaviors without explicit finger pose supervision. Through experiments across multiple datasets and hand morphologies, we show that wrist-guided learning can achieve grasping and object manipulation performance comparable to methods that rely on detailed finger supervision. These results suggest that emphasizing wrist guidance may offer a practical direction for reducing reliance on dense finger motion capture in physics-based manipulation learning. More broadly, this perspective suggests that manipulation policies can be learned by focusing on structurally important joints rather than densely supervising all degrees of freedom.

\section*{Acknowledgements}

This work was supported by Samsung Electronics (IO201208-07822-01) and also by IITP grants (RS-2022-II220113: Developing a Sustainable Collaborative Multi-modal Lifelong Learning Framework (50\%), RS-2022-II220290: Visual Intelligence for Space-Time Understanding \& Generation (25\%), RS-2024-00457882: National AI Research Lab Project (20\%), RS-2019-II191906: AI Graduate School Program at POSTECH (5\%)) funded by Ministry of Science and ICT, Korea.

\newpage
\bibliographystyle{splncs04}
\bibliography{main}

@String(TOG= {ACM Trans. Graph.})

@String(TOG   = {ACM TOG})

@article{ppo,
  title={Proximal policy optimization algorithms},
  author={Schulman, John and Wolski, Filip and Dhariwal, Prafulla and Radford, Alec and Klimov, Oleg},
  journal={arXiv preprint arXiv:1707.06347},
  year={2017}
}

@inproceedings{smplx,
  title={Expressive body capture: 3d hands, face, and body from a single image},
  author={Pavlakos, Georgios and Choutas, Vasileios and Ghorbani, Nima and Bolkart, Timo and Osman, Ahmed AA and Tzionas, Dimitrios and Black, Michael J},
  booktitle={Proceedings of the IEEE/CVF conference on computer vision and pattern recognition},
  pages={10975--10985},
  year={2019}
}

@article{deepmimic,
  title={Deepmimic: Example-guided deep reinforcement learning of physics-based character skills},
  author={Peng, Xue Bin and Abbeel, Pieter and Levine, Sergey and Van de Panne, Michiel},
  journal={ACM Transactions On Graphics (TOG)},
  volume={37},
  number={4},
  pages={1--14},
  year={2018},
  publisher={ACM New York, NY, USA}
}

@article{amp,
  title={AMP: Adversarial motion priors for stylized physics-based character control},
  author={Peng, Xue Bin and Ma, Ze and Abbeel, Pieter and Levine, Sergey and Kanazawa, Angjoo},
  journal={ACM Transactions on Graphics (TOG)},
  volume={40},
  number={4},
  pages={1--20},
  year={2021},
  publisher={ACM}
}

@inproceedings{parahome,
  title={Parahome: Parameterizing everyday home activities towards 3d generative modeling of human-object interactions},
  author={Kim, Jeonghwan and Kim, Jisoo and Na, Jeonghyeon and Joo, Hanbyul},
  booktitle={Proceedings of the Computer Vision and Pattern Recognition Conference},
  pages={1816--1828},
  year={2025}
}

@article{ase,
  title={Ase: Large-scale reusable adversarial skill embeddings for physically simulated characters},
  author={Peng, Xue Bin and Guo, Yunrong and Halper, Lina and Levine, Sergey and Fidler, Sanja},
  journal={ACM Transactions On Graphics (TOG)},
  volume={41},
  number={4},
  pages={1--17},
  year={2022},
  publisher={ACM New York, NY, USA}
}

@article{pulse,
  title={Universal humanoid motion representations for physics-based control},
  author={Luo, Zhengyi and Cao, Jinkun and Merel, Josh and Winkler, Alexander and Huang, Jing and Kitani, Kris and Xu, Weipeng},
  journal={arXiv preprint arXiv:2310.04582},
  year={2023}
}

@inproceedings{phc,
  title={Perpetual humanoid control for real-time simulated avatars},
  author={Luo, Zhengyi and Cao, Jinkun and Kitani, Kris and Xu, Weipeng and others},
  booktitle={Proceedings of the IEEE/CVF International Conference on Computer Vision},
  pages={10895--10904},
  year={2023}
}

@inproceedings{intermimic,
  title={Intermimic: Towards universal whole-body control for physics-based human-object interactions},
  author={Xu, Sirui and Ling, Hung Yu and Wang, Yu-Xiong and Gui, Liang-Yan},
  booktitle={Proceedings of the Computer Vision and Pattern Recognition Conference},
  pages={12266--12277},
  year={2025}
}

@inproceedings{skillmimic,
  title={Skillmimic: Learning basketball interaction skills from demonstrations},
  author={Wang, Yinhuai and Zhao, Qihan and Yu, Runyi and Tsui, Hok Wai and Zeng, Ailing and Lin, Jing and Luo, Zhengyi and Yu, Jiwen and Li, Xiu and Chen, Qifeng and others},
  booktitle={Proceedings of the Computer Vision and Pattern Recognition Conference},
  pages={17540--17549},
  year={2025}
}

@inproceedings{interphys,
  title={Synthesizing physical character-scene interactions},
  author={Hassan, Mohamed and Guo, Yunrong and Wang, Tingwu and Black, Michael and Fidler, Sanja and Peng, Xue Bin},
  booktitle={ACM SIGGRAPH 2023 Conference Proceedings},
  pages={1--9},
  year={2023}
}

@inproceedings{grab,
  title={GRAB: A dataset of whole-body human grasping of objects},
  author={Taheri, Omid and Ghorbani, Nima and Black, Michael J and Tzionas, Dimitrios},
  booktitle={European conference on computer vision},
  pages={581--600},
  year={2020},
  organization={Springer}
}

@inproceedings{amass,
  title={AMASS: Archive of motion capture as surface shapes},
  author={Mahmood, Naureen and Ghorbani, Nima and Troje, Nikolaus F and Pons-Moll, Gerard and Black, Michael J},
  booktitle={Proceedings of the IEEE/CVF international conference on computer vision},
  pages={5442--5451},
  year={2019}
}

@article{maskedmanipulator,
  title={MaskedManipulator: Versatile Whole-Body Control for Loco-Manipulation},
  author={Tessler, Chen and Jiang, Yifeng and Coumans, Erwin and Luo, Zhengyi and Chechik, Gal and Peng, Xue Bin},
  journal={arXiv preprint arXiv:2505.19086},
  year={2025}
}

@inproceedings{skillmimicv2,
  title={Skillmimic-v2: Learning robust and generalizable interaction skills from sparse and noisy demonstrations},
  author={Yu, Runyi and Wang, Yinhuai and Zhao, Qihan and Tsui, Hok Wai and Wang, Jingbo and Tan, Ping and Chen, Qifeng},
  booktitle={Proceedings of the Special Interest Group on Computer Graphics and Interactive Techniques Conference Conference Papers},
  pages={1--11},
  year={2025}
}

@inproceedings{dexgraspnet,
  title={DexGraspNet: A large-scale robotic dexterous grasp dataset for general objects based on simulation},
  author={Wang, Ruicheng and Zhang, Jialiang and Chen, Jiayi and Xu, Yinzhen and Li, Puhao and Liu, Tengyu and Wang, He},
  booktitle={IEEE International Conference on Robotics and Automation (ICRA)},
  pages={11359--11366},
  year={2023}
}

@inproceedings{oakink,
  title={Oakink: A large-scale knowledge repository for understanding hand-object interaction},
  author={Yang, Lixin and Li, Kailin and Zhan, Xinyu and Wu, Fei and Xu, Anran and Liu, Liu and Lu, Cewu},
  booktitle={Proceedings of the IEEE/CVF conference on computer vision and pattern recognition},
  pages={20953--20962},
  year={2022}
}

@inproceedings{oakink2,
  title={Oakink2: A dataset of bimanual hands-object manipulation in complex task completion},
  author={Zhan, Xinyu and Yang, Lixin and Zhao, Yifei and Mao, Kangrui and Xu, Hanlin and Lin, Zenan and Li, Kailin and Lu, Cewu},
  booktitle={Proceedings of the IEEE/CVF Conference on Computer Vision and Pattern Recognition},
  pages={445--456},
  year={2024}
}

@inproceedings{unidexgrasp,
  title={Unidexgrasp: Universal robotic dexterous grasping via learning diverse proposal generation and goal-conditioned policy},
  author={Xu, Yinzhen and Wan, Weikang and Zhang, Jialiang and Liu, Haoran and Shan, Zikang and Shen, Hao and Wang, Ruicheng and Geng, Haoran and Weng, Yijia and Chen, Jiayi and others},
  booktitle={Proceedings of the IEEE/CVF Conference on Computer Vision and Pattern Recognition},
  pages={4737--4746},
  year={2023}
}

@article{omnigrasp,
  title={Omnigrasp: Grasping diverse objects with simulated humanoids},
  author={Luo, Zhengyi and Cao, Jinkun and Christen, Sammy and Winkler, Alexander and Kitani, Kris and Xu, Weipeng},
  journal={Advances in Neural Information Processing Systems},
  volume={37},
  pages={2161--2184},
  year={2024}
}

@article{dexmachina,
  title={Dexmachina: Functional retargeting for bimanual dexterous manipulation},
  author={Mandi, Zhao and Hou, Yifan and Fox, Dieter and Narang, Yashraj and Mandlekar, Ajay and Song, Shuran},
  journal={arXiv preprint arXiv:2505.24853},
  year={2025}
}

@article{dexvlg,
  title={DexVLG: Dexterous Vision-Language-Grasp Model at Scale},
  author={He, Jiawei and Li, Danshi and Yu, Xinqiang and Qi, Zekun and Zhang, Wenyao and Chen, Jiayi and Zhang, Zhaoxiang and Zhang, Zhizheng and Yi, Li and Wang, He},
  journal={arXiv preprint arXiv:2507.02747},
  year={2025}
}

@article{dexgraspvla,
  title={Dexgraspvla: A vision-language-action framework towards general dexterous grasping},
  author={Zhong, Yifan and Huang, Xuchuan and Li, Ruochong and Zhang, Ceyao and Chen, Zhang and Guan, Tianrui and Zeng, Fanlian and Lui, Ka Num and Ye, Yuyao and Liang, Yitao and others},
  journal={arXiv preprint arXiv:2502.20900},
  year={2025}
}

@inproceedings{physhoi,
  title={Physically plausible full-body hand-object interaction synthesis},
  author={Braun, Jona and Christen, Sammy and Kocabas, Muhammed and Aksan, Emre and Hilliges, Otmar},
  booktitle={2024 International Conference on 3D Vision (3DV)},
  pages={464--473},
  year={2024},
  organization={IEEE}
}

@inproceedings{gym,
  author    = {Viktor Makoviychuk and Lukasz Wawrzyniak and Yunrong Guo and Michelle Lu and Kier Storey and M. Macklin and David Hoeller and N. Rudin and Arthur Allshire and Ankur Handa and Gavriel State},
  year      = {2021},
  title     = {Isaac Gym: High Performance GPU-Based Physics Simulation For Robot Learning},
  booktitle = {NeurIPS Datasets and Benchmarks},
}

@ARTICLE{silu,
  title         = "Gaussian Error Linear Units ({GELUs})",
  author        = "Hendrycks, Dan and Gimpel, Kevin",
  month         =  jun,
  year          =  2016,
  copyright     = "http://arxiv.org/licenses/nonexclusive-distrib/1.0/",
  archivePrefix = "arXiv",
  primaryClass  = "cs.LG",
  eprint        = "1606.08415"
}

@Inproceedings{PMP,
 author = {Jinseok Bae and Jungdam Won and Donggeun Lim and Cheol-Hui Min and Y. Kim},
 booktitle = {International Conference on Computer Graphics and Interactive Techniques},
 title = {PMP: Learning to Physically Interact with Environments using Part-wise Motion Priors},
 year = {2023}
}

@inproceedings{hot3d,
  title={Hot3d: Hand and object tracking in 3d from egocentric multi-view videos},
  author={Banerjee, Prithviraj and Shkodrani, Sindi and Moulon, Pierre and Hampali, Shreyas and Han, Shangchen and Zhang, Fan and Zhang, Linguang and Fountain, Jade and Miller, Edward and Basol, Selen and others},
  booktitle={Proceedings of the Computer Vision and Pattern Recognition Conference},
  pages={7061--7071},
  year={2025}
}

@inproceedings{arctic,
  title={ARCTIC: A dataset for dexterous bimanual hand-object manipulation},
  author={Fan, Zicong and Taheri, Omid and Tzionas, Dimitrios and Kocabas, Muhammed and Kaufmann, Manuel and Black, Michael J and Hilliges, Otmar},
  booktitle={Proceedings of the IEEE/CVF conference on computer vision and pattern recognition},
  pages={12943--12954},
  year={2023}
}

@inproceedings{freihand,
  title={Freihand: A dataset for markerless capture of hand pose and shape from single rgb images},
  author={Zimmermann, Christian and Ceylan, Duygu and Yang, Jimei and Russell, Bryan and Argus, Max and Brox, Thomas},
  booktitle={Proceedings of the IEEE/CVF international conference on computer vision},
  pages={813--822},
  year={2019}
}

@inproceedings{ho3d,
  title={Honnotate: A method for 3d annotation of hand and object poses},
  author={Hampali, Shreyas and Rad, Mahdi and Oberweger, Markus and Lepetit, Vincent},
  booktitle={Proceedings of the IEEE/CVF conference on computer vision and pattern recognition},
  pages={3196--3206},
  year={2020}
}

@inproceedings{h2o,
  title={H2o: Two hands manipulating objects for first person interaction recognition},
  author={Kwon, Taein and Tekin, Bugra and St{\"u}hmer, Jan and Bogo, Federica and Pollefeys, Marc},
  booktitle={Proceedings of the IEEE/CVF international conference on computer vision},
  pages={10138--10148},
  year={2021}
}

@inproceedings{dexycb,
  title={DexYCB: A benchmark for capturing hand grasping of objects},
  author={Chao, Yu-Wei and Yang, Wei and Xiang, Yu and Molchanov, Pavlo and Handa, Ankur and Tremblay, Jonathan and Narang, Yashraj S and Van Wyk, Karl and Iqbal, Umar and Birchfield, Stan and others},
  booktitle={Proceedings of the IEEE/CVF conference on computer vision and pattern recognition},
  pages={9044--9053},
  year={2021}
}

@inproceedings{assemblyhands,
  title={Assemblyhands: Towards egocentric activity understanding via 3d hand pose estimation},
  author={Ohkawa, Takehiko and He, Kun and Sener, Fadime and Hodan, Tomas and Tran, Luan and Keskin, Cem},
  booktitle={Proceedings of the IEEE/CVF conference on computer vision and pattern recognition},
  pages={12999--13008},
  year={2023}
}

@article{tactile,
  title={Tactile sensing—from humans to humanoids},
  author={Dahiya, Ravinder S and Metta, Giorgio and Valle, Maurizio and Sandini, Giulio},
  journal={IEEE transactions on robotics},
  volume={26},
  number={1},
  pages={1--20},
  year={2009},
  publisher={IEEE}
}

@article{3dvitac,
  title={3d-vitac: Learning fine-grained manipulation with visuo-tactile sensing},
  author={Huang, Binghao and Wang, Yixuan and Yang, Xinyi and Luo, Yiyue and Li, Yunzhu},
  journal={arXiv preprint arXiv:2410.24091},
  year={2024}
}

@article{difftactile,
  title={Difftactile: A physics-based differentiable tactile simulator for contact-rich robotic manipulation},
  author={Si, Zilin and Zhang, Gu and Ben, Qingwei and Romero, Branden and Xian, Zhou and Liu, Chao and Gan, Chuang},
  journal={arXiv preprint arXiv:2403.08716},
  year={2024}
}

@article{mimictouch,
  title={Mimictouch: Leveraging multi-modal human tactile demonstrations for contact-rich manipulation},
  author={Yu, Kelin and Han, Yunhai and Wang, Qixian and Saxena, Vaibhav and Xu, Danfei and Zhao, Ye},
  journal={arXiv preprint arXiv:2310.16917},
  year={2023}
}

@article{omnih2o,
  title={Omnih2o: Universal and dexterous human-to-humanoid whole-body teleoperation and learning},
  author={He, Tairan and Luo, Zhengyi and He, Xialin and Xiao, Wenli and Zhang, Chong and Zhang, Weinan and Kitani, Kris and Liu, Changliu and Shi, Guanya},
  journal={arXiv preprint arXiv:2406.08858},
  year={2024}
}

@article{okami,
  title={Okami: Teaching humanoid robots manipulation skills through single video imitation},
  author={Li, Jinhan and Zhu, Yifeng and Xie, Yuqi and Jiang, Zhenyu and Seo, Mingyo and Pavlakos, Georgios and Zhu, Yuke},
  journal={arXiv preprint arXiv:2410.11792},
  year={2024}
}

@article{expressive,
  title={Expressive whole-body control for humanoid robots},
  author={Cheng, Xuxin and Ji, Yandong and Chen, Junming and Yang, Ruihan and Yang, Ge and Wang, Xiaolong},
  journal={arXiv preprint arXiv:2402.16796},
  year={2024}
}

@article{hermes,
  title={Hermes: Human-to-robot embodied learning from multi-source motion data for mobile dexterous manipulation},
  author={Yuan, Zhecheng and Wei, Tianming and Gu, Langzhe and Hua, Pu and Liang, Tianhai and Chen, Yuanpei and Xu, Huazhe},
  journal={arXiv preprint arXiv:2508.20085},
  year={2025}
}

@inproceedings{vividex,
  title={Vividex: Learning vision-based dexterous manipulation from human videos},
  author={Chen, Zerui and Chen, Shizhe and Arlaud, Etienne and Laptev, Ivan and Schmid, Cordelia},
  booktitle={2025 IEEE International Conference on Robotics and Automation (ICRA)},
  pages={3336--3343},
  year={2025},
  organization={IEEE}
}

@article{humanplus,
  title={Humanplus: Humanoid shadowing and imitation from humans},
  author={Fu, Zipeng and Zhao, Qingqing and Wu, Qi and Wetzstein, Gordon and Finn, Chelsea},
  journal={arXiv preprint arXiv:2406.10454},
  year={2024}
}

@inproceedings{interhand2,
  title={Interhand2. 6m: A dataset and baseline for 3d interacting hand pose estimation from a single rgb image},
  author={Moon, Gyeongsik and Yu, Shoou-I and Wen, He and Shiratori, Takaaki and Lee, Kyoung Mu},
  booktitle={European Conference on Computer Vision},
  pages={548--564},
  year={2020},
  organization={Springer}
}

@inproceedings{meng20223d,
  title={3d interacting hand pose estimation by hand de-occlusion and removal},
  author={Meng, Hao and Jin, Sheng and Liu, Wentao and Qian, Chen and Lin, Mengxiang and Ouyang, Wanli and Luo, Ping},
  booktitle={European Conference on Computer Vision},
  pages={380--397},
  year={2022},
  organization={Springer}
}

@article{omomo,
  title={Object motion guided human motion synthesis},
  author={Li, Jiaman and Wu, Jiajun and Liu, C Karen},
  journal={ACM Transactions on Graphics (TOG)},
  volume={42},
  number={6},
  pages={1--11},
  year={2023},
  publisher={ACM New York, NY, USA}
}

@article{
    scadiver,
    author = {Won, Jungdam and Gopinath, Deepak and Hodgins, Jessica},
    title = {A Scalable Approach to Control Diverse Behaviors for Physically Simulated Characters},
    year = {2020},
    issue_date = {July 2020},
    volume = {39},
    number = {4},
    url = {https://doi.org/10.1145/3386569.3392381},
    journal = {ACM Trans. Graph.},
    articleno = {33},
}

@article{gmt,
  title={Gmt: General motion tracking for humanoid whole-body control},
  author={Chen, Zixuan and Ji, Mazeyu and Cheng, Xuxin and Peng, Xuanbin and Peng, Xue Bin and Wang, Xiaolong},
  journal={arXiv preprint arXiv:2506.14770},
  year={2025}
}

@article{beyondmimic,
  title={Beyondmimic: From motion tracking to versatile humanoid control via guided diffusion},
  author={Liao, Qiayuan and Truong, Takara E and Huang, Xiaoyu and Gao, Yuman and Tevet, Guy and Sreenath, Koushil and Liu, C Karen},
  journal={arXiv preprint arXiv:2508.08241},
  year={2025}
}

@article{any2track,
  title={Track any motions under any disturbances},
  author={Zhang, Zhikai and Guo, Jun and Chen, Chao and Wang, Jilong and Lin, Chenghuai and Lian, Yunrui and Xue, Han and Wang, Zhenrong and Liu, Maoqi and Lyu, Jiangran and others},
  journal={arXiv preprint arXiv:2509.13833},
  year={2025}
}

@article{unitracker,
  title={Unitracker: Learning universal whole-body motion tracker for humanoid robots},
  author={Yin, Kangning and Zeng, Weishuai and Fan, Ke and Dai, Minyue and Wang, Zirui and Zhang, Qiang and Tian, Zheng and Wang, Jingbo and Pang, Jiangmiao and Zhang, Weinan},
  journal={arXiv preprint arXiv:2507.07356},
  year={2025}
}

@article{sonic,
  title={Sonic: Supersizing motion tracking for natural humanoid whole-body control},
  author={Luo, Zhengyi and Yuan, Ye and Wang, Tingwu and Li, Chenran and Chen, Sirui and Castaneda, Fernando and Cao, Zi-Ang and Li, Jiefeng and Minor, David and Ben, Qingwei and others},
  journal={arXiv preprint arXiv:2511.07820},
  year={2025}
}

@inproceedings{tokenhsi,
  title={Tokenhsi: Unified synthesis of physical human-scene interactions through task tokenization},
  author={Pan, Liang and Yang, Zeshi and Dou, Zhiyang and Wang, Wenjia and Huang, Buzhen and Dai, Bo and Komura, Taku and Wang, Jingbo},
  booktitle={Proceedings of the Computer Vision and Pattern Recognition Conference},
  pages={5379--5391},
  year={2025}
}

@inproceedings{hoifhli,
  title={Human-object interaction from human-level instructions},
  author={Wu, Zhen and Li, Jiaman and Xu, Pei and Liu, C Karen},
  booktitle={Proceedings of the IEEE/CVF International Conference on Computer Vision},
  pages={11176--11186},
  year={2025}
}

@article{closd,
  title={Closd: Closing the loop between simulation and diffusion for multi-task character control},
  author={Tevet, Guy and Raab, Sigal and Cohan, Setareh and Reda, Daniele and Luo, Zhengyi and Peng, Xue Bin and Bermano, Amit H and van de Panne, Michiel},
  journal={arXiv preprint arXiv:2410.03441},
  year={2024}
}

@inproceedings{maniptrans,
  title={Maniptrans: Efficient dexterous bimanual manipulation transfer via residual learning},
  author={Li, Kailin and Li, Puhao and Liu, Tengyu and Li, Yuyang and Huang, Siyuan},
  booktitle={Proceedings of the IEEE/CVF Conference on Computer Vision and Pattern Recognition},
  pages={6991--7003},
  year={2025}
}

@article{spider,
  title={Spider: Scalable physics-informed dexterous retargeting},
  author={Pan, Chaoyi and Wang, Changhao and Qi, Haozhi and Liu, Zixi and Bharadhwaj, Homanga and Sharma, Akash and Wu, Tingfan and Shi, Guanya and Malik, Jitendra and Hogan, Francois},
  journal={arXiv preprint arXiv:2511.09484},
  year={2025}
}

@inproceedings{graspd,
  title={Grasp’d: Differentiable contact-rich grasp synthesis for multi-fingered hands},
  author={Turpin, Dylan and Wang, Liquan and Heiden, Eric and Chen, Yun-Chun and Macklin, Miles and Tsogkas, Stavros and Dickinson, Sven and Garg, Animesh},
  booktitle={European Conference on Computer Vision},
  pages={201--221},
  year={2022},
  organization={Springer}
}

@article{dexdiffuser,
  title={Dexdiffuser: Generating dexterous grasps with diffusion models},
  author={Weng, Zehang and Lu, Haofei and Kragic, Danica and Lundell, Jens},
  journal={IEEE Robotics and Automation Letters},
  volume={9},
  number={12},
  pages={11834--11840},
  year={2024},
  publisher={IEEE}
}

@inproceedings{ugg,
  title={Ugg: Unified generative grasping},
  author={Lu, Jiaxin and Kang, Hao and Li, Haoxiang and Liu, Bo and Yang, Yiding and Huang, Qixing and Hua, Gang},
  booktitle={European Conference on Computer Vision},
  pages={414--433},
  year={2024},
  organization={Springer}
}

@inproceedings{dgtr,
  title={Dexterous grasp transformer},
  author={Xu, Guo-Hao and Wei, Yi-Lin and Zheng, Dian and Wu, Xiao-Ming and Zheng, Wei-Shi},
  booktitle={Proceedings of the IEEE/CVF Conference on Computer Vision and Pattern Recognition},
  pages={17933--17942},
  year={2024}
}

@article{dexgys,
  title={Grasp as you say: Language-guided dexterous grasp generation},
  author={Wei, Yi-Lin and Jiang, Jian-Jian and Xing, Chengyi and Tan, Xian-Tuo and Wu, Xiao-Ming and Li, Hao and Cutkosky, Mark and Zheng, Wei-Shi},
  journal={Advances in Neural Information Processing Systems},
  volume={37},
  pages={46881--46907},
  year={2024}
}

@inproceedings{semgrasp,
  title={Semgrasp: Semantic grasp generation via language aligned discretization},
  author={Li, Kailin and Wang, Jingbo and Yang, Lixin and Lu, Cewu and Dai, Bo},
  booktitle={European Conference on Computer Vision},
  pages={109--127},
  year={2024},
  organization={Springer}
}

@article{dexonomy,
  title={Dexonomy: Synthesizing all dexterous grasp types in a grasp taxonomy},
  author={Chen, Jiayi and Ke, Yubin and Peng, Lin and Wang, He},
  journal={arXiv preprint arXiv:2504.18829},
  year={2025}
}

@inproceedings{afforddexgrasp,
  title={Afforddexgrasp: Open-set language-guided dexterous grasp with generalizable-instructive affordance},
  author={Wei, Yi-Lin and Lin, Mu and Lin, Yuhao and Jiang, Jian-Jian and Wu, Xiao-Ming and Zeng, Ling-An and Zheng, Wei-Shi},
  booktitle={Proceedings of the IEEE/CVF International Conference on Computer Vision},
  pages={11818--11828},
  year={2025}
}

@article{dexter,
  title={DextER: Language-driven Dexterous Grasp Generation with Embodied Reasoning},
  author={Lee, Junha and Park, Eunha and Cho, Minsu},
  journal={arXiv preprint arXiv:2601.16046},
  year={2026}
}

@article{rubiks,
  title={Solving rubik's cube with a robot hand},
  author={Akkaya, Ilge and Andrychowicz, Marcin and Chociej, Maciek and Litwin, Mateusz and McGrew, Bob and Petron, Arthur and Paino, Alex and Plappert, Matthias and Powell, Glenn and Ribas, Raphael and others},
  journal={arXiv preprint arXiv:1910.07113},
  year={2019}
}

@article{andrychowicz2020learning,
  title={Learning dexterous in-hand manipulation},
  author={Andrychowicz, OpenAI: Marcin and Baker, Bowen and Chociej, Maciek and Jozefowicz, Rafal and McGrew, Bob and Pachocki, Jakub and Petron, Arthur and Plappert, Matthias and Powell, Glenn and Ray, Alex and others},
  journal={The International Journal of Robotics Research},
  volume={39},
  number={1},
  pages={3--20},
  year={2020},
  publisher={SAGE Publications Sage UK: London, England}
}

@article{qin2022one,
  title={From one hand to multiple hands: Imitation learning for dexterous manipulation from single-camera teleoperation},
  author={Qin, Yuzhe and Su, Hao and Wang, Xiaolong},
  journal={IEEE Robotics and Automation Letters},
  volume={7},
  number={4},
  pages={10873--10881},
  year={2022},
  publisher={IEEE}
}

@inproceedings{egomimic,
  title={Egomimic: Scaling imitation learning via egocentric video},
  author={Kareer, Simar and Patel, Dhruv and Punamiya, Ryan and Mathur, Pranay and Cheng, Shuo and Wang, Chen and Hoffman, Judy and Xu, Danfei},
  booktitle={2025 IEEE International Conference on Robotics and Automation (ICRA)},
  pages={13226--13233},
  year={2025},
  organization={IEEE}
}

@inproceedings{dextreme,
  title={Dextreme: Transfer of agile in-hand manipulation from simulation to reality},
  author={Handa, Ankur and Allshire, Arthur and Makoviychuk, Viktor and Petrenko, Aleksei and Singh, Ritvik and Liu, Jingzhou and Makoviichuk, Denys and Van Wyk, Karl and Zhurkevich, Alexander and Sundaralingam, Balakumar and others},
  booktitle={2023 IEEE International Conference on Robotics and Automation (ICRA)},
  pages={5977--5984},
  year={2023},
  organization={IEEE}
}

@article{chen2023sequential,
  title={Sequential dexterity: Chaining dexterous policies for long-horizon manipulation},
  author={Chen, Yuanpei and Wang, Chen and Fei-Fei, Li and Liu, C Karen},
  journal={arXiv preprint arXiv:2309.00987},
  year={2023}
}

@inproceedings{rotateit,
  title={General in-hand object rotation with vision and touch},
  author={Qi, Haozhi and Yi, Brent and Suresh, Sudharshan and Lambeta, Mike and Ma, Yi and Calandra, Roberto and Malik, Jitendra},
  booktitle={Conference on Robot Learning},
  pages={2549--2564},
  year={2023},
  organization={PMLR}
}

@article{dexumi,
  title={Dexumi: Using human hand as the universal manipulation interface for dexterous manipulation},
  author={Xu, Mengda and Zhang, Han and Hou, Yifan and Xu, Zhenjia and Fan, Linxi and Veloso, Manuela and Song, Shuran},
  journal={arXiv preprint arXiv:2505.21864},
  year={2025}
}

@article{egoscale,
  title={EgoScale: Scaling Dexterous Manipulation with Diverse Egocentric Human Data},
  author={Zheng, Ruijie and Niu, Dantong and Xie, Yuqi and Wang, Jing and Xu, Mengda and Jiang, Yunfan and Casta{\~n}eda, Fernando and Hu, Fengyuan and Tan, You Liang and Fu, Letian and others},
  journal={arXiv preprint arXiv:2602.16710},
  year={2026}
}

@article{chen2024object,
  title={Object-centric dexterous manipulation from human motion data},
  author={Chen, Yuanpei and Wang, Chen and Yang, Yaodong and Liu, C Karen},
  journal={arXiv preprint arXiv:2411.04005},
  year={2024}
}

@inproceedings{graspxl,
  title={Graspxl: Generating grasping motions for diverse objects at scale},
  author={Zhang, Hui and Christen, Sammy and Fan, Zicong and Hilliges, Otmar and Song, Jie},
  booktitle={European Conference on Computer Vision},
  pages={386--403},
  year={2024},
  organization={Springer}
}

\clearpage
\begin{center}
    {\Large\bfseries Supplementary Material\par}
\end{center}
\vspace{1.2em}

\appendix
\setcounter{table}{0}
\renewcommand{\thetable}{\Alph{table}}
\renewcommand*{\theHtable}{\thetable}
\setcounter{figure}{0}
\renewcommand{\thefigure}{\Alph{figure}}
\renewcommand*{\theHfigure}{\thefigure}

\section{Additional Results}
\label{sec:supp_additional_results}

\subsection{Ablation on Finger Guidance}
\label{subsec:finger_guidance}
\begin{table}[h]
    \centering
    \vspace{-5mm}
    \setlength{\tabcolsep}{4pt}
    \resizebox{\linewidth}{!}{%
    \begin{tabular}{lccc}
        \toprule
         & \multicolumn{3}{c}{ParaHome} \\
        \cmidrule(lr){2-4}
         Method & Success rate (\%) $\uparrow$ & Obj. pos. err. (cm) $\downarrow$ & Obj. rot. err. ($^\circ$) $\downarrow$ \\
        \midrule
        Decoupled (No Wrist Constraints) & 0.0 & 46.5 & 42.9 \\
        WristMimic (Ours) & \textbf{83.3} & \textbf{15.3} & \textbf{33.9} \\
        Ours + Finger Guidance & 68.7 & 29.2 & 38.9 \\
        \bottomrule
    \end{tabular}
    }
    \vspace{2mm}
    \caption{Quantitative comparison on ParaHome~\cite{parahome}. Best results are indicated in \textbf{bold}.}
    \label{tab:supp_parahome_quantitative}
    \vspace{-10mm}
\end{table}

\Tblstart{supp_parahome_quantitative} reports results on the extended 20-sequence ParaHome split, complementing the 8-sequence ablation in Tab. 2 of the main paper and adding a variant with explicit finger-tracking guidance. WristMimic achieves superior performance across all three metrics compared to the decoupled baseline without wrist constraints (Decoupled). Interestingly, WristMimic which does not use explicit finger kinematic trajectories in either the state or reward formulation, also outperforms the variant with additional finger tracking guidance. This observation suggests that prioritizing object kinematics and contact outcomes can provide a stronger learning signal than directly supervising high-dimensional finger poses. One possible explanation is the extremely high degrees of freedom of the hands (45 DoFs per hand), where enforcing finger pose tracking may introduce large pose deviations or unintended wrist misalignment while satisfying finger-related rewards. As a result, focusing on wrist placement together with object-centric objectives leads to more stable and effective manipulation behavior.

\subsection{Ablation on Embodiment Generalization}
\label{subsec:supp_hand_embodiment}
\begin{figure}[t]
    \centering
    \includegraphics[width=0.7\linewidth]{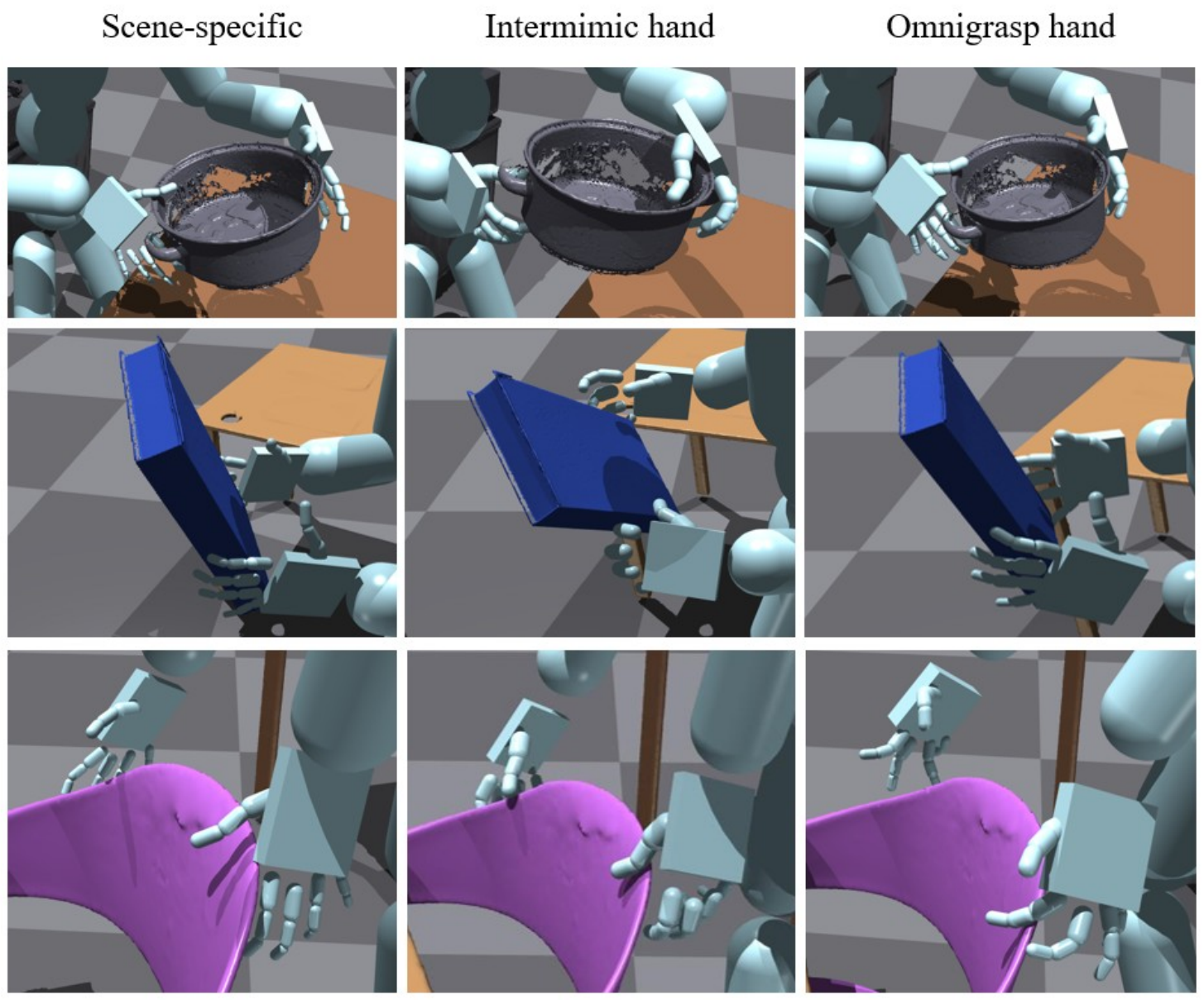}
    \vspace{-2mm}
    \caption{Additional qualitative results on hand embodiment generalization.}
    \label{fig:supp_hand_embodiment}
    \vspace{-5mm}
\end{figure}

\Figstart{supp_hand_embodiment} presents additional experimental results showing robots with different hand morphologies grasping a variety of objects, supplementing Fig.~6 in the main paper. WristMimic consistently produces stable grasp configurations across diverse hand embodiments, demonstrating that the learned manipulation strategy remains robust despite variations in hand geometry.

\begin{figure}[t]
    \centering
    \includegraphics[width=0.7\linewidth]{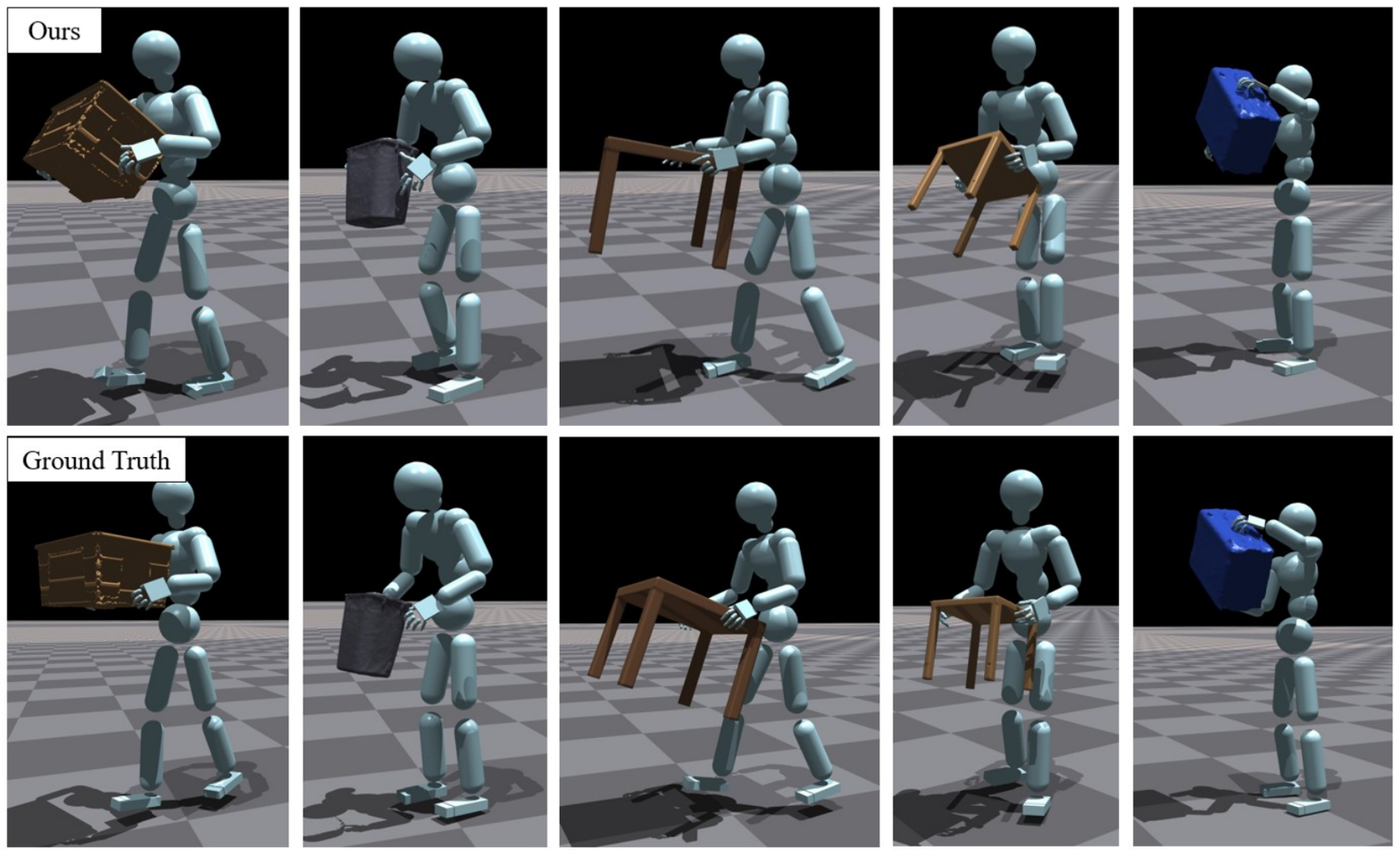}
    \vspace{-6mm}
    \caption{Qualitative results on OMOMO~\cite{omomo} dataset.}
    \label{fig:supp_omomo_qual}
\end{figure}
\subsection{Qualitative Results  on OMOMO dataset}
\label{subsec:supp_omomo_qual}
\Figstart{supp_omomo_qual} demonstrates qualitative results of our method on OMOMO~\cite{omomo} dataset. In contrast to the reference motion exhibiting unstable grasping and floating hand artifacts, our method produces accurate grasping poses that are faithful to object dynamics.

\section{Simulation Setup}

\subsection{Physical Assets}

\noindent\textbf{Humanoid.} We use scene-specific SMPL-X~\cite{smplx} to generate humanoid bodies following OMOMO~\cite{omomo} and InterMimic~\cite{intermimic}. Physical parameters in ~\Tbl{humanoid_params} use higher stiffness for torso and spine (1000) for stability and lower stiffness for fingers (100) for compliant grasping.

\begin{table}[h]
\centering
\vspace{-5mm}
\resizebox{0.7\columnwidth}{!}{
\begin{tabular}{lcccc}
\toprule
\textbf{Body Part} & \textbf{Stiffness} & \textbf{Damping} & \textbf{Armature} & \textbf{Density (kg/m\textsuperscript{3})} \\
\midrule
Leg & 800 & 80 & 0.02 & 1000 \\
Torso and Spine & 1000 & 100 & 0.02 & 1000 \\
Arm & 500 & 50 & 0.02 & 1000 \\
Fingers & 100 & 10 & 0.02 & 1000 \\
Toe & 500 & 50 & 0.02 & 1000 \\
\bottomrule
\end{tabular}
}
\vspace{2mm}
\caption{Humanoid physical parameters. All body parts use the same density (1000 kg/m\textsuperscript{3}) and armature (0.02). Joint ranges follow InterMimic~\cite{intermimic} for all hands}
\label{tab:humanoid_params}
\vspace{-10mm}
\end{table}

\noindent\textbf{Objects.} Static objects in ParaHome(table, desk, gasstove, bookshelf, sink) have fixed base links. Dynamic objects use provided meshes as visual geometry with collision meshes generated by uniformly sampling 20\% vertices. Parameters in ~\Tbl{object_params} show that we set max\_convex\_hulls to 80 for hollow-handled objects (kettle, cup), 50 for structurally complex furniture (chair), and 10 for simpler objects.

\begin{table}[h]
\centering
\resizebox{0.8\columnwidth}{!}{
\begin{tabular}{lll}
\toprule
\textbf{Parameter} & \textbf{Value} & \textbf{Special Cases} \\
\midrule
Angular damping & 0.01 & -- \\
Linear damping & 0.01 & -- \\
Default DOF drive mode & DOF\_MODE\_NONE & -- \\
Max convex hulls & 10 & kettle, cup: 80; chair: 50 \\
Max vertices per convex hull & 64 & -- \\
VHACD resolution & 300000 & -- \\
\bottomrule
\end{tabular}
}
\vspace{2mm}
\caption{Object asset parameters for Isaac Gym~\cite{gym}. Static objects (e.g., table, desk, gasstove, bookshelf, sink) have fixed base links. Dynamic objects use visual mesh as-is and collision mesh generated by 20\% uniform vertex sampling.}
\vspace{-5mm}
\label{tab:object_params}
\end{table}

\subsection{Simulator Configuration}

We use Isaac Gym~\cite{gym} with PhysX backend. Key parameters in ~\Tbl{simulator_config} include num\_position\_iterations set to 10 (vs. default 4) to prevent object penetration during grasping, which is critical for stable contact-rich manipulation.

\section{Training Details}

\noindent\textbf{Training configuration.} We train with 2048 parallel environments using minibatch size 8192 and horizon length 32. Details are in ~\Tbl{training_config}.

\begin{table*}[h!]
\centering
\scriptsize
\begin{minipage}[t]{0.50\textwidth}
\centering
\resizebox{0.95\linewidth}{!}{
\begin{tabular}{ll}
\toprule
\textbf{Parameter} & \textbf{Value} \\
\midrule
Substeps & 2 \\
Control frequency & 30Hz \\
Num threads & 4 \\
Solver type & 1 (TGS) \\
Num position iterations & 10 \\
Num velocity iterations & 0 \\
Contact offset (m) & 0.002 \\
Rest offset (m) & 0.0 \\
Bounce threshold velocity (m/s) & 0.2 \\
Max depenetration velocity (m/s) & 10.0 \\
Default buffer size multiplier & 20.0 \\
Max GPU contact pairs & 34603008 \\
\bottomrule
\end{tabular}
}
\vspace{2mm}
\captionof{table}{Isaac Gym~\cite{gym} simulator settings with PhysX backend.}
\label{tab:simulator_config}
\end{minipage}
\hfill
\begin{minipage}[t]{0.45\textwidth}
\centering
\resizebox{1.0\linewidth}{!}{
\begin{tabular}{ll}
\toprule
\textbf{Parameter} & \textbf{Value} \\
\midrule
Contact window before ($t_b$) & 10 frames \\
Contact window after ($t_a$) & 15 frames \\
Num envs & 2048 \\
Minibatch size & 8192 \\
Horizon length & 32 \\
GPU & RTX3090 \\
\bottomrule
\end{tabular}
}
\vspace{2mm}
\captionof{table}{Training configuration and rollout parameters.}
\label{tab:training_config}
\end{minipage}
\end{table*}

\noindent\textbf{Reward weights.} Our reward function follows the multiplicative form $r = \exp(-\lambda E)$, where $E$ is the tracking error and $\lambda$ is the reward weight. See \Tbl{reward_weights} for detailed reward weights.

\begin{table}[h!]
\centering
\resizebox{0.5\columnwidth}{!}{
\begin{tabular}{lrl}
\toprule
\textbf{Component} & \textbf{Weight} & \textbf{Description} \\
\midrule
p & 30.0 & Body joint position \\
r & 2.5 & Body joint rotation \\
op & 5.0 & Object position \\
or & 0.1 & Object rotation \\
ig & 5.0 & Interaction graph \\
cg\_hand & 10.0 & Hand contact \\
cg\_other & 5.0 & Other body contact \\
\midrule
gbp & 10.0 & Body position \\
gbr & 1.0 & Body rotation \\
gwp & 10.0 & Wrist position \\
gwr & 1.0 & Wrist rotation \\
\midrule
eg1 & 2e-5 & Body joint energy \\
eg2 & 2e-5 & Object energy \\
eg3 & 1e-9 & Contact energy \\
\bottomrule
\end{tabular}
}
\vspace{2mm}
\caption{Reward weights. Upper arm joints set to zero during contact window. Velocity rewards (pv, rv, opv, orv) excluded.}
\label{tab:reward_weights}
\end{table}

\noindent\textbf{Design rationale.} During the contact window, we reduce body weights (gbp: 30$\rightarrow$10, gbr: 2.5$\rightarrow$1) and zero upper arm joints to focus on contact dynamics. Wrist weights stay constant (gwp = 10, gwr = 1) across all phases, with stage-specific reset boundaries providing the primary guidance. The grasping phase uses strict thresholds (7cm position, 0.2 rad orientation), while approach and stabilization phases allow more exploration (15cm, 0.5 rad).

\end{document}